\documentclass[10pt,journal,compsoc]{IEEEtran}
\usepackage{graphicx}
\usepackage{amsmath}
\usepackage{amssymb}
\usepackage{subfigure}
\usepackage{times}
\usepackage{latexsym}
\usepackage{microtype}
\usepackage{color}
\usepackage{float}
\usepackage{tabularx,multirow}
\usepackage{hyperref}
\usepackage[hyphenbreaks]{breakurl}
\makeatletter
\def\UrlAlphabet{%
      \do\a\do\b\do\c\do\d\do\e\do\f\do\g\do\h\do\i\do\j%
      \do\k\do\l\do\m\do\n\do\o\do\p\do\q\do\r\do\s\do\t%
      \do\u\do\v\do\w\do\x\do\y\do\z\do\A\do\B\do\C\do\D%
      \do\E\do\F\do\G\do\H\do\I\do\J\do\K\do\L\do\M\do\N%
      \do\O\do\P\do\Q\do\R\do\S\do\T\do\U\do\V\do\W\do\X%
      \do\Y\do\Z}
\def\UrlDigits{\do\1\do\2\do\3\do\4\do\5\do\6\do\7\do\8\do\9\do\0}
\g@addto@macro{\UrlBreaks}{\UrlOrds}
\g@addto@macro{\UrlBreaks}{\UrlAlphabet}
\g@addto@macro{\UrlBreaks}{\UrlDigits}

\usepackage{makecell}
\usepackage{booktabs}

\usepackage{bm}

\usepackage{ulem}

\ifCLASSOPTIONcompsoc
  \usepackage[nocompress]{cite}
\else
  \usepackage{cite}
\fi
\ifCLASSINFOpdf
\else
\fi
\begin{document}
%
\title{Inconsistent Matters: A Knowledge-guided Dual-consistency Network for Multi-modal Rumor Detection}
%
%
%
%
\author{Mengzhu~Sun,
        Xi~Zhang,
        Jianqiang Ma, Sihong Xie,
        Yazheng Liu,
        and Philip S. Yu~\IEEEmembership{Fellow,~IEEE,}
\IEEEcompsocitemizethanks{\IEEEcompsocthanksitem Mengzhu Sun, Xi Zhang and Yazheng Liu are with Beijing University of Posts and Telecommunications, Beijing 100876, China. \protect E-mail: \{2019110945, zhangx, liuyz\}@bupt.edu.cn.

\IEEEcompsocthanksitem Jianqiang Ma is with the Platform and Content Group, Tencent, Beijing 100080, China. \protect E-mail: alexanderma@tencent.com.

\IEEEcompsocthanksitem Sihong Xie is with the Lehigh University, Bethlehem, PA 18015, USA. \protect E-mail: six316@lehigh.edu.

\IEEEcompsocthanksitem Philip S. Yu is with the University
of Illinois at Chicago, Chicago, IL 60607, USA. \protect E-mail: psyu@uic.edu.}

\thanks{Manuscript received April 19, 2005; revised August 26, 2015.\\(Corresponding author: Xi Zhang)}}

%
%

\markboth{Journal of \LaTeX\ Class Files,~Vol.~14, No.~8, August~2015}%
{Shell \MakeLowercase{\textit{et al.}}: Bare Demo of IEEEtran.cls for Computer Society Journals}
%



\IEEEtitleabstractindextext{%
\begin{abstract}
Rumor spreaders are increasingly utilizing multimedia content to attract the attention and trust of news consumers. Though quite a few rumor detection models have exploited the multi-modal data, they seldom consider the inconsistent semantics between images and texts, and rarely spot the inconsistency among the post contents and background knowledge. In addition, they commonly assume the completeness of multiple modalities and thus are incapable of handling handle missing modalities in real-life scenarios. Motivated by the intuition that rumors in social media are more likely to have inconsistent semantics, a novel \emph{Knowledge-guided Dual-consistency Network} is proposed to detect rumors with multimedia contents. It uses two consistency detection subnetworks to capture the inconsistency at the cross-modal level and the content-knowledge level simultaneously. It also enables robust multi-modal representation learning under different missing visual modality conditions, using a special token to discriminate between posts with visual modality and posts without visual modality.
Extensive experiments on three public real-world multimedia datasets demonstrate that our framework can outperform the state-of-the-art baselines under both complete and incomplete modality conditions. Our codes are available at \url{https://github.com/MengzSun/KDCN}.

\end{abstract}

\begin{IEEEkeywords}
Rumor Detection, Multi-modal Learning, Social Media Analysis.
\end{IEEEkeywords}}

\maketitle

\IEEEdisplaynontitleabstractindextext

%
\IEEEpeerreviewmaketitle

\IEEEraisesectionheading{\section{Introduction}\label{sec:introduction}}

%
%
%
%
\IEEEPARstart{T}{he} rapid growth of social media has revolutionized the way people acquire news. Unfortunately, social media has fostered various false information, including misrepresented or even forged multimedia content, to mislead readers. The widespread rumors may cause significant adverse effects. For example, some offenders use rumors to manipulate public opinion, damage the credibility of the government, and even interfere with the general election ~\cite{allcott2017social}. Therefore, it is urgent to automatically detect and regulate rumors to promote trust in the social media ecosystem.

Traditional rumor detection methods mainly rely on textual data to extract distinctive features~\cite{castillo2011information,chen2018call,10.5555/3061053.3061153,10.5555/3172077.3172434}. With the advancement of multimedia technology, visual contents have become an important part of rumors to attract and mislead the consumers due to more credible storytelling and rapid diffusion~\cite{jin2016novel,qi2019exploiting}. To this end, the rumor detection methods are undergoing a transition from a uni-modal to a multi-modal paradigm. 

\begin{figure*}[htbp]
    \centering
    \subfigure[One real-world example of a fake multimedia tweet to show cross-modal inconsistency. Its textual content ``the two suspected \#CharlieHebdo gunmen have been killed." has nothing to do with its image content that something behind the woods is on fire. ]{\includegraphics[height=7.5cm,width=0.45\linewidth]{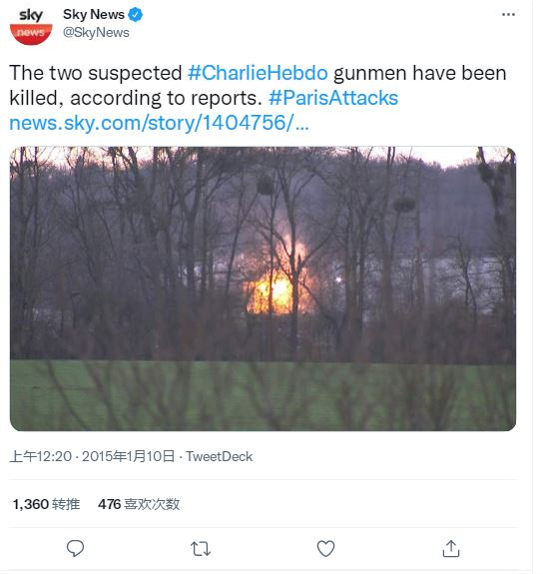}}\hspace{5mm}
    \subfigure[The other real-world example of a fake multimedia tweet to show content-knowledge inconsistency. It is suspicious to see sharks appear in a subway. Such abnormality should be captured and serve as an essential clue for rumor identification.]{\includegraphics[height=7.5cm,width=0.45\linewidth]{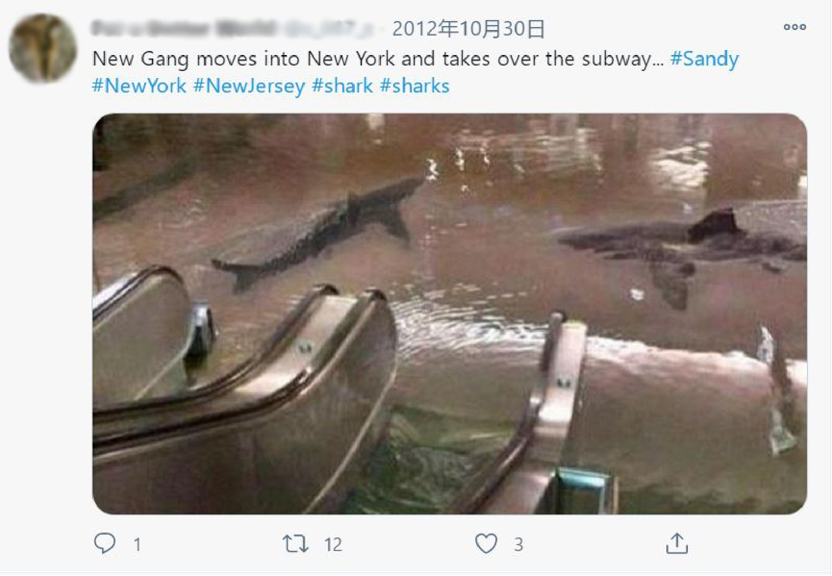}}
    \caption{Two real-world examples of fake multi-modal tweets.}
    \label{fig1_example}
\end{figure*}

Detecting multimedia rumor posts is a double-edged sword. On the one hand, it is more challenging to learn effective feature representations from heterogeneous multi-modal data. On the other hand, it also provides a great opportunity to identify inconsistent cues among multi-modal data. Xue et al. \cite{xue2021detecting} show that to catch the eyes of the public, rumors tend to use theatrical, comical, and attractive images that are irrelevant to the post content. In general, it is often difficult to find pertinent and non-manipulated images to match fictional events. And thus posts with mismatched textual and visual information are very likely to be fake~\cite{zhou2020mathsf}. Fig.~\ref{fig1_example} (a) shows a real-world multimedia rumor from Twitter, where there is a fire somewhere in the image that has nothing to do with the textual content ``two gunmen have been killed''. Thus, it is essential to identify such \emph{cross-modal inconsistency} for multimedia rumor identification. Additionally, one major drawback of these multi-modal methods is that they assume the availability of paired data modalities in both training and testing data. However, in many real-world scenarios, not all modalities are available. For example, a large number of posts on Twitter or Weibo have only textual contents, without the visual modality. Compared with discarding any data points with missing modality in previous studies~\cite{wang2018eann,khattar2019mvae,9376933,zhou2020mathsf}, including these data points may lead to more representativeness of the training data and thus better generalizability to the test data, which is one major issue we aim to solve.

In addition to using visual information, rumor detection can also benefit from the introduction of knowledge graphs (KG), which can provide faithful background knowledge to verify the semantic integrity of post contents. 
Previous works~\cite{zhang2019multi,wang2020fake} commonly used KG to complement the post contents by various data fusion methods. However, they ignore the \emph{content-knowledge inconsistency} information. For example, in Fig.~\ref{fig1_example} (b), it would be a great help to judge the truthfulness of the post, given the background knowledge that sharks are unlikely to appear in a subway. 
Intuitively, if we are able to spot the uncommon co-occurring entities in the multi-modal post contents, such as the entity pair ``shark" and ``subway" in Fig.\ref{fig1_example} (b)~\footnote{Note that entity inconsistency is not necessarily cross-modal as shown in this example.}, it would facilitate the detection of counterintuitive rumors.


Although a few recent multi-modal rumor detection methods have captured the image-text dissimilarity as an indicative feature, they fail to consider the \emph{content-knowledge inconsistency} at the same time. The two types of consistency information can complement each other, so that even if one is unreliable (for example, no text-image dissimilarity is detected in Fig.\ref{fig1_example} (b)), the other can help. Also, the two types of information can have some complex interactions that can be learned by a deep network to discover more efficient detection signals. Thus, it would be beneficial to exploit both types of information for better rumor detection.

Along this line, in this work, we aim to exploit both \emph{cross-modal inconsistency} and \emph{content-knowledge inconsistency} for multimedia rumor detection, without requiring full modalities. The problem is non-trivial due to the following challenges. First, since text, image, and KG data have different formats and structures, how to integrate them into a unified framework to detect rumors is an open question. Second, there is no straightforward way to measure and capture the aforementioned
inconsistency.
Third, an effective detector is expected to robustly adapt to different visual modality missing patterns: modality missing in training data, testing data, or both. 

To address the above challenges, we propose a novel \emph{Knowledge-guided Dual-Consistency Network (KDCN)} that can capture the inconsistent information at the cross-modal level and the content-knowledge level simultaneously. To validate our motivation that inconsistency matters for rumor detection, we analyze the rumor datasets and observe that the above two types of inconsistency information present a statistically significant distinction between rumor and non-rumor posts (see details in Sec. \ref{experiments:preliminary}). Following this observation, our framework mainly consists of two sub-neural networks: one is to extract cross-modal differences between images and texts, and the other is to identify the abnormal co-occurrence of pairs of entities in the post contents by measuring their KG representation distances. The two sub-neural networks are tightly coupled to make the two sources of inconsistency information complement each other, which can improve the robustness of the detection of rumors, even if one source is unavailable or unreliable. Moreover, to enable our framework to tackle the incomplete modalities, we utilize pseudo images as a complement with a special token to indicate it is not real. It is simple and can make our framework unaltered to process the incomplete modality data with the same procedure as modality-complete data, and meanwhile provide stable performance under different cases of missing visual modality.  
To summarize, the contributions of our paper are three-fold:
 \begin{itemize}
\item We propose a novel knowledge-guided dual-consistency network to simultaneously capture the cross-modal inconsistency and content-knowledge inconsistency. It is designed to detect rumors with multi-modal contents, but can also adapt to cases where the visual modality is missing.


\item  To the best of our knowledge, 
we are the \textit{first} to reveal that rumor posts tend to contain entities that are farther away on KG than non-rumors.
This observation can serve as a useful signal to distinguish between rumors and non-rumors. 
\item Extensive experiments on three real-world datasets show that our framework can better detect rumors than the state-of-the-art baselines. It is also advantageous in providing stable and robust performance under different visual modality missing patterns, even under very severe missing scenarios.

\end{itemize} 
\section{Related Work} 

\label{sec:length}

\subsection{Rumor Detection}


Rumor detection models rely on various features extracted from multi-modal social media data, including post texts, social context,  the attached images, and the related knowledge graphs. Thus, we review existing work from the following four categories: textual and social contextual-based methods, multimedia methods, fact-checking with KG, and knowledge-enhanced methods. 

\subsubsection{Textual and social contextual rumor detection} 
Most rumor detection models rely on \emph{textual features}. Traditional machine learning-based models are based on features extracted from textual posts in a feature engineering manner~\cite{zhao2015enquiring,castillo2011information}. Recent studies propose deep learning models to capture high-level textual semantics, outperforming traditional machine learning-based models. A recurrent neural network (RNN) based model is proposed to capture the variation of contextual information of relevant posts over time  ~\cite{10.5555/3061053.3061153}. ~\cite{ma-etal-2018-rumor} proposes a user-attention-based convolutional neural network (CNN) model with an adversarial cross-lingual learning framework to capture both the language-specific and language-independent features. ~\cite{10.5555/3172077.3172434} proposes a convolutional approach for misinformation identification based on CNN to extract key textual features. ~\cite{9802916} proposes multi-channel networks to model news pieces from semantic, emotional, and stylistic views.

\emph{Social context features} represent the user engagements on social media such as retweeting and commenting behaviors. Social context features can provide important clues to differentiate rumors from non-rumors. ~\cite{shu2019defend} develops a sentence-comment co-attention sub-network to exploit both news contents and user comments to jointly capture important sentences and user comments as explanations to support the detection result. ~\cite{tian2020qsan} proposes a quantum-probability-based signed attention network utilizing post contents and related comments to detect false information. Both of these two studies utilize retweeting and commenting content. ~\cite{8939421} proposes a repost-based early rumor detection model by regarding all reposts of a post as a sequence.
~\cite{wu2015false} proposes a graph-kernel based hybrid SVM classifier to capture the high-order propagation patterns. This study uses network structures as social context features. However, social context features are 
usually unavailable at the early stage of news dissemination.



\subsubsection{Multimedia rumor detection} 
Several recent models begin to explore the role of visual information. \cite{jin2017multimodal} proposes a recurrent neural network to extract and fuse multi-modal and social context features with an attention mechanism. EANN~\cite{wang2018eann} learns post representations by leveraging both the textual and visual information, using an adversarial method to remove event-specific features to benefit newly arrived events. \cite{khattar2019mvae} proposes a multi-modal variational autoencoder for rumor detection to learn shared features from both modalities. The encoder encodes the information from text and image into a latent vector, while the decoder reconstructs the original image and text. \cite{9376933} designs a multi-modal multi-task learning framework by introducing the stance task. However, these studies do not consider consistencies between multi-modal information as our work does. While SAFE~\cite{zhou2020mathsf} and MCNN~\cite{xue2021detecting} have considered the relevance between textual and visual information, we distance our work from theirs in that we capture the cross-modal inconsistency differently, and also model the inconsistency between content and external knowledge. In addition, these studies don't touch the modality missing issue, which is common for real-world multi-modal rumor detection. COSMOS~\cite{DBLP:journals/corr/abs-2101-06278} focuses on a new task of predicting whether the image has been used out of context by taking as input an image and two corresponding captions from two different news sources. If the two captions refer to the same object in the image, but are semantically different, then it indicates out-of-context use of image. It has a different problem setting from this work.



\subsubsection{Fact-checking with KG} 
Some studies~\cite{ciampaglia2015computational,fionda2018fact,pan2018content,shi2016discriminative} extract structured triples (head, relation, tail) from the post contents, and fact-check them with the faithful triples in KG. A limitation of such approaches is that KG is typically incomplete or imprecise to cover the complex relations in the form of triple being extracted from the post.  
Consider an extracted triple (Anthony Weiner, cooperate with, FBI) has two entities with the ``cooperate with" relation, where both entities are available in KG, but the relation is not~\cite{pan2018content}. For such cases, structured triple methods fail to make reliable predictions. 
By contrast, our method is still applicable.



\subsubsection{Knowledge-enhanced detection} 
A few studies use external knowledge to supplement post contents to obtain better representations for rumor detection. A knowledge-guided article embedding is learned for healthcare misinformation detection by incorporating medical knowledge graph and propagating the node embeddings through knowledge paths~\cite{cui2020deterrent}. The multi-modal knowledge-aware representation and event-invariant features are learned together to form the event representation in ~\cite{zhang2019multi}, which is fed into a deep neural network for rumor detection. A knowledge-driven multi-modal graph convolutional network (KMGCN)~\cite{wang2020fake} is proposed to model the global structure among texts, images, and knowledge concepts to obtain comprehensive semantic representations. ~\cite{10.1145/3474085.3481548} proposes an entity-enhanced multi-modal fusion framework, which models correlations of entity inconsistency, mutual enhancement, and text complementation to detect multi-modal rumors. ~\cite{hu-etal-2021-compare} proposes a graph neural model, which compares the news to the knowledge base (KB) through entities for fake news detection. However, these methods don't consider the content-knowledge inconsistency. Moreover, KMGCN is transductive, requiring the inferred nodes to be present at training time, and is time-consuming due to graph construction and learning.





\subsection{Multi-modal Learning with Missing Modality}\label{related:missing modality}


Modalities can be partially missing in multi-modal learning tasks. For example, due to lighting or occlusion issues, faces can not always be detected for the emotion recognition task~\cite{zhao2021missing}, resulting in modality missing. One solution to this problem is data augmentation, where missing modality cases are simulated by randomly ablating the inputs ~\cite{parthasarathy2020training}. Another common solution is using generative methods. Given the available modalities, the missing modalities are predicted directly~\cite{li2018video,cai2018deep,suo2019metric,du2018semi}. Some studies learn joint multi-modal representations from these modalities~\cite{aguilar-etal-2019-multimodal,pham2019found,han2019implicit,wang2020transmodality,zhao2021missing}.    

Note that most of the existing methods are designed for the scenario that full modalities do exist but cannot be accessed due to various constraints. However, for the rumor detection task, the visual modality is missing mostly since there don't exist any corresponding images at all. Therefore, the previous approaches such as generative methods may incur unnecessary computational cost and bring large noises. To the best of our knowledge, how to tackle the incompleteness of images for multi-modal rumor detection has not been covered by existing studies. Moreover, due to the large number of posts on social media, a lightweight way is expected to provide superior and robust performance for different missing cases.

\begin{figure*}[htbp]
    \centering
    \includegraphics[width=0.9\linewidth]{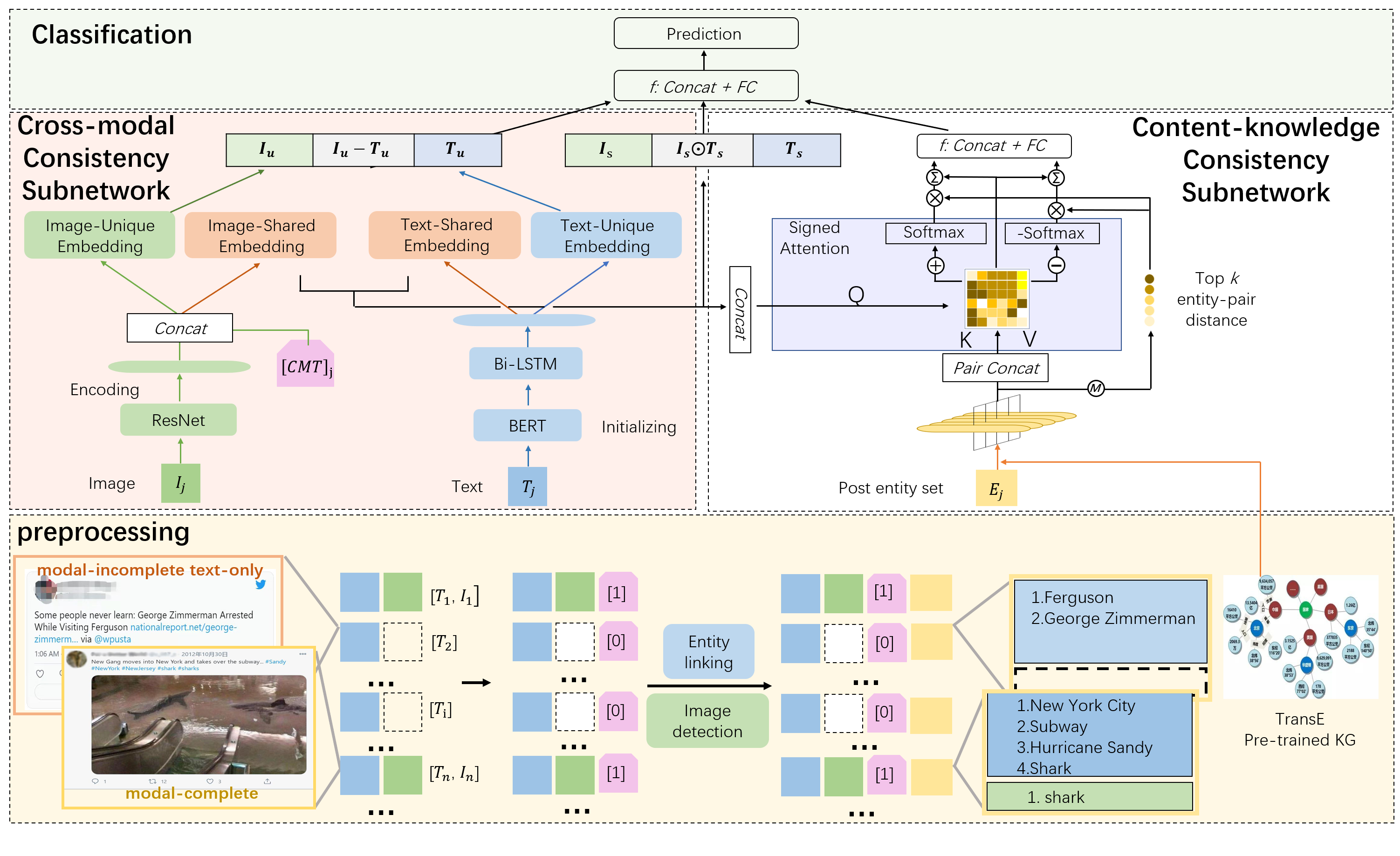}
    \caption{The framework of the proposed knowledge-guided dual-consistency network. It consists of four components: (1) bottom: \emph{the data preprocessing component}. For the text-only post, a pseudo image (represented by a white square) is used to fill the position of the missing visual data, and a token [CMT] = 0 is used to represent a text-only post (represented by a pink hexagon). For a post from the modal-complete dataset, a token [CMT] = 1 is used to represent a post with an image. This component extracts and links the entity mentions from multimedia contents to the corresponding entities in KG. A post entity set is represented by a yellow square. Then the entities are represented with pre-trained embeddings; (2) middle left: \emph{the cross-modal consistency subnetwork}. It encodes the image and text, and the CMT token is concatenated to the image representation. Then, it projects them into modal-shared and modal-unique spaces, and learns the cross-modal inconsistency features. (3) middle right: \emph{the content-knowledge consistency subnetwork}. For a post entity set, an entity pair representation EP is formed by concatenating any two entities from the set. In the figure, this operation is represented as \emph{Pair Concat}. The Manhattan distances are calculated between any two entities from the set, and we get the top-$k$ entity pairs with the largest Manhattan distances and their corresponding distances. This operation is represented as \emph{M}. This component uses the modal-shared content as query Q and the entity pair representations EP as the value and key, and a distance-aware signed attention mechanism that adopts both ``+Softmax'' and ``-Softmax'' operations to capture multi-aspect correlations to obtain content-knowledge inconsistency features as in Eq. (8) and (9); (4) top: \emph{the rumor classification layer} to combine the cross-modal inconsistency features, modal-shared features and content-knowledge inconsistency features. \emph{Concat} denotes the concatenation operation, and \emph{FC} represents the fully-connected layer.
} \label{fig:2 framework}
\end{figure*}

\section{Methodology}

\subsection{Problem Definition}
Following previous studies~\cite{wang2018eann,zhou2020mathsf,khattar2019mvae}, the rumor detection task can be defined as a binary classification problem with the two classes of rumor or non-rumor.
In this paper, without loss of generality, we consider a multi-modal rumor dataset involving the visual and textual modalities, where some samples may lack the visual modality. Formally, let $\mathcal{D} = \{\mathcal{D}^f, \mathcal{D}^t\}$ denote the overall \emph{modal-incomplete dataset}, and all posts in $\mathcal{D}$ can be divided into two subsets $\mathcal{D}^f$ and $\mathcal{D}^t$ according to the presence or absence of the visual modal data, respectively.
$\mathcal{D}^f = \{T_i, I_i, y_i \}_i$ denotes the \emph{modal-complete subset}, where $T_i$ represents the textual data and $I_i$ represents the visual data of the $i$-th sample. $y_i$ is the corresponding class label.
$\mathcal{D}^t = \{T_j, y_j \}_j$ denotes the \emph{text-only subset}, where the visual data is missing.
Our goal is to leverage both modal-complete and text-only subsets for model training. The proposed model needs to adapt to different visual-modality missing conditions, that is, the visual data can be missing in the training data, testing data, or both.


\subsection{Overview}


As shown in Fig~.\ref{fig:2 framework}, our framework mainly consists of four components : (1) a \emph{preprocessig component} to obtain entities and their representations; (2) a \emph{cross-modal consistency subnetwork} for capturing the inconsistency between image and text for each post. This subnetwork also has to deal with the visual modality missing issue; (3) a \emph{content-knowledge consistency subnetwork} for capturing the inconsistency between the content and KG through entity distances; (4) a \emph{classification layer} that aggregates various features and produces classification labels. 


The data flow is as follows. Given a social post from dataset $\mathcal{D}$, this post can have both textual and visual modalities, or have textual modality only. We first extract entities from texts (and images, if the visual modality is also available) and obtain the entity representations. The collection of entity representations is fed into the content-knowledge consistency subnetwork to get the knowledge-level inconsistency features. Meanwhile, for a specific post, a special token [CMT] is introduced as an indicator to determine whether this post belongs to the \emph{modal-complete subset} $\mathcal{D}^f$ or the \emph{text-only subset} $\mathcal{D}^t$. If the post belongs to the \emph{text-only subset}, since it lacks visual data, we supplement the post with a pseudo image to make it compatible with the cross-modal consistency subnetwork. Then the image and text data, as well as the token are fed into the cross-modal consistency subnetwork to produce cross-modal inconsistency features and modal-shared features. After going through the above two consistency subnetworks, the obtained features are fused and fed into the classification layer to produce final labels. In the following sections, we will describe each component in detail.

\subsection{Multi-modal Post Preprocessing}\label{method:preprocessing}
For the posts in the modal-complete subset $\mathcal{D}^f$, we essentially follow the procedure in~\cite{wang2020fake} to extract entities from texts and images. For the text content, we use the entity linking solution TAGME\footnote{TAGME is available at \url{https://tagme.d4science.org/tagme/}}~\cite{ASSANTE2019555} and Shuyantech\footnote{Shuyantech is available at \url{http://shuyantech.com/entitylinking}}~\cite{chen2018short} to extract and link the ambiguous entity mentions in the text to the corresponding entities in KG for English and Chinese texts, respectively. For the visual content, we utilize the off-the-shelf pre-trained YOLOv3\footnote{YOLOv3 pre-trained model is provided in \url{https://pjreddie.com/darknet/yolo/\#demo}}~\cite{yolov3} to extract semantic objects as visual words. The labels of detected objects, such as person and dog, are treated as entity mentions. These mentions are linked to entities in KG.

Then, the entity in the text modality is linked to entities in KG. In this paper, we take Freebase\footnote{Freebase data dumps is available at \url{https://developers.google.com/freebase/}} as the reference KG. The reasons why we choose Freebase as the knowledge source are two-fold: (1) Freebase has a much larger scale set of entities than Probase and Yago, which would facilitate the rumor detection task. (2) There are off-the-shelf pre-trained entity embeddings that can be used directly by our model. We then obtain the pre-trained entity representations from the publicly available OpenKE\footnote{OpenKE is available at \url{http://openke.thunlp.org}} , which are trained with TransE~\cite{bordes2013translating} on Freebase. The entity representation embedding dimension is 50. Thus, our model accepts
quadruple inputs $\{$Text, Image, Entity~set, Pretrained~KG$\}$. How to process the data instances without the visual modality would be illustrated in Sec.~\ref{method:cross-modal:missing}.



\subsection{Cross-modal Consistency Subnetwork}\label{method:cross-modal}

The cross-modal consistency subnetwork is designed to capture the inconsistency between images and texts and deal with the visual modality missing issue. It consists of two separate encoders for texts and images, a decomposition layer to obtain the corresponding modal-unique features and modal-shared features, and a fusion layer to produce cross-modal inconsistency features. 


\subsubsection{Text and image encoding}\label{method:cross-modal:encoding}
We map texts and images into feature representations. Specifically, for the text information, we use the initial word embeddings pre-trained by BERT, and utilize the bi-directional long short-term memory (Bi-LSTM) network to encode each textual sequence into a vector following the procedure in~\cite{9317523}.
In particular, it maps the word embedding \bm{$w_j$} into its hidden state \bm{$h_j$}$\in \mathbb{R}^{d_0}$,
where \bm{$w_j$}$\in \mathbb{R}^{d_w}$ denotes the pre-trained  embedding of the $j$-th word from a word sequence with length $M$. We concatenate \bm{$\overleftarrow{h_0}$} and \bm{$\overrightarrow{h_M}$} to obtain the hidden state of the textual content \bm{$h$} $\in \mathbb{R}^{2d_0}$. After that, we encode the textual representation into a $d$-dimensional vector \bm{$H_T$},

\begin{equation}
    \begin{aligned}
        \boldsymbol{H_T} = {\rm ReLU}(\boldsymbol{w_T}*\boldsymbol{h}+\boldsymbol{b_T}),
    \end{aligned}
\end{equation}
where $\boldsymbol{w_T} \in \mathbb{R}^{d \times 2d_o}$ and $\boldsymbol{b_T} \in \mathbb{R}^{d \times 1}$ are learnable weights and bias parameters.


Similarly, we encode an image into a $d$-dimensional vector \bm{$\hat{H_I}$} with a pre-trained CNN,



\begin{equation}
    \begin{aligned}
        \boldsymbol{\hat{H_I}} = {\rm ReLU}(\boldsymbol{\hat{w_I}}*(\mathbf{CNN}(Image)+\boldsymbol{\hat{b_I}}),
    \end{aligned}
\end{equation}
where $\boldsymbol{\hat{w_I}}$ $\in \mathbb{R}^{d \times d_I}$ and $\boldsymbol{\hat{b_I}}$ $\in \mathbb{R}^{d \times 1}$ are learnable parameters, $d_I$ is the dimension of the pre-trained CNN image vector. However, here we assume the visual data is available.  How to make it compatible with those posts where the visual modality data is missing would be introduced in the following part.


\subsubsection{Pseudo image for visual modality missing}\label{method:cross-modal:missing}

Till now, we have assumed full modality data are available for multi-modal data preprocessing and encoding. We then discuss how to process the data instances where the visual modality data is missing. 

As stated in Sec.~\ref{related:missing modality}, one common solution to address the  missing modality issue is to use generative methods. But they are designed for the scenario that full modalities do exist but cannot be accessed due to various constraints. However, for the rumor detection task, it is common that the visual modality does not exist in the source post, and thus it is not necessary to generate the images at all. Moreover, generating images based on the available textual modality would incur heavy computational costs in handling the large number of posts on the social network.

To address this issue, we propose a novel approach that uses a pseudo image with a special token to supplement these data instances. By taking this approach, we can address the problem of the incompleteness of modalities in terms of flexibility (missing modalities in training, testing, or both) without alternating the framework architecture. It is also advantageous in efficiency as no extra training or generative overhead is required. Moreover, different from traditional methods that discard the data instances with missing modality, it can take full advantage of the training data and can thus better generalize to the test data.


Specifically, for each post in the text-only subset $D^t=\{T_j, y_j\}_j$, the text modality is processed in the same way as the modal-complete post described in Sec.~\ref{method:cross-modal:encoding}. To address the visual data missing issue, we propose to fill the position of the visual data with a pseudo image. Concretely, we use a white image (RGB(255, 255, 255 ) as the pseudo visual data. To distinguish it from the real image, a special Complete-Modality Token ([CMT]) is introduced. [CMT]=\{0,1\}, where 0 indicates that the post is from the text-only subset, and 1 indicates coming from the modal-complete subset.

After that, our model accepts quintuple inputs: $\{$Text, Image, Entity~set, Pretrained~KG, [CMT] = 1$\}$ for the modal-complete subset $\mathcal{D}^f$ and $\{$Text, pseudo~Image, Entity~set, Pretrained~KG, [CMT]=0$\}$ for the text-only subset $\mathcal{D}^t$.

Then we improve the image encoding method in Eq.~(2) to make it accommodate both real and pseudo images. Specifically, we put the corresponding complete-modality token [CMT] after every image representation. They are concatenated and mapped into a low $d$ -dimension space:
\begin{equation}
    \begin{aligned}
        \boldsymbol{H_I} = {\rm ReLU}(\boldsymbol{w_I}*[\mathbf{CNN}(Image);[{\rm CMT}]]+\boldsymbol{b_I}),
    \end{aligned}
\end{equation}
where $\boldsymbol{w_I}$ $\in \mathbb{R}^{d \times (d_I+1)}$ and $\boldsymbol{b_I}$ $\in \mathbb{R}^{d \times 1}$ are learnable parameters. The effect of [CMT] will be verified in the experimental section.

Please note that besides the above [CMT] token method, we have also tried to generate images based on generative adversarial networks as well as use randomly generated images to serve as the missing images. The performance of these comparison methods is reported in Sec. 4.6.

\subsubsection{Multi-modal decomposition} Enlightened by the idea of projecting the multi-modal representations into different spaces~\cite{xu-etal-2020-reasoning}, we break down the raw visual and textual representations into the modal-unique space and modal-shared space. While a cross-modal shared layer is proposed to extract modal-invariant shared features, an image-specific layer and a text-specific layer are used to extract the corresponding modal-unique features:
\begin{equation}\label{method:cross-modal:3}
    \begin{aligned}
        \boldsymbol{I_{s}} &= \boldsymbol{W_{shared}}\boldsymbol{H_I} \in \mathbb{R}^{d_s}\\  \boldsymbol{I_{u}} &= \boldsymbol{P_{I}}\boldsymbol{H_I} \in \mathbb{R}^{d_u}\\
        \boldsymbol{T_{s}} &= \boldsymbol{W_{shared}}\boldsymbol{H_T} \in \mathbb{R}^{d_s}\\ \boldsymbol{T_{u}} &= \boldsymbol{P_{T}}\boldsymbol{H_T} \in \mathbb{R}^{d_u}
    \end{aligned}
\end{equation}
where \bm{$H_I$} and \bm{$H_T$} are the encoded visual and textual features obtained in the last subsection, \bm{$W_{shared}$} $\in \mathbb{R}^{d_s \times d}$ and $\{\boldsymbol{P_I}, \boldsymbol{P_T}\} \in \mathbb{R}^{d_u \times d}$ are projection matrices for the modal-shared space and modal-unique space, respectively. \bm{$I_s$} and \bm{$I_u$} are the decomposed modal-shared and modal-unique image features, respectively, while \bm{$T_s$} and \bm{$T_u$} are the decomposed modal-shared and modal-unique text features, respectively.

To ensure that the decomposed modal-shared space is unrelated with the modal-unique spaces, the orthogonal constrain is introduced as:
\begin{equation}
\begin{aligned}
    \boldsymbol{W_{shared}} (\bm{P_I})^{T} = 0\\
    \boldsymbol{W_{shared}} (\bm{P_T})^{T} = 0
\end{aligned}
\end{equation}
which can be converted into the following orthogonal loss, 
\begin{equation}\label{method:cross-modal:5}
    \mathcal{L}_o = ||\boldsymbol{W_{shared}} (\boldsymbol{P_I})^{T}||_F^2 + ||\boldsymbol{W_{shared}} (\boldsymbol{P_T})^{T}||_F^2,
\end{equation}
where $||\cdot||_F^2$ denotes the Forbenius norm. We verify that the orthogonal loss is useful for improving detection performance in the ablation study in Sec. \ref{experiments:ablation}.

After obtaining two modal-unique features and two modal-shared features in Eq. \ref{method:cross-modal:3}, we combine them as the cross-modal inconsistency representation \bm{$f_{unique}$} and the overall modal-shared representation \bm{$f_{share}$}, that is
\begin{equation}
    \begin{aligned}
        \boldsymbol{f_{unique}} &= [\boldsymbol{T_u};\boldsymbol{T_u}-\boldsymbol{I_u};\boldsymbol{I_u}]\\
        \boldsymbol{f_{share}} &= [\boldsymbol{T_s};\boldsymbol{T_s} \odot \boldsymbol{I_s};\boldsymbol{I_s}],
    \end{aligned}
\end{equation}
where $\odot$ denotes the element-wise multiplication operation, \bm{$f_{unique}$} $\in \mathbb{R}^{3d_u}$ is used to measure the inconsistency information between modalities, and \bm{$f_{share}$} $\in \mathbb{R}^{3d_s}$ is used to represent the shared information between modalities. 
Similar ideas to obtain the cross-modal contrast features can also be found in~\cite{xu-etal-2020-reasoning}. But unlike it which only focuses on the opposition between different modalities, we also retain the modal-shared content to preserve the comprehensive multi-modal semantics. Then both \bm{$f_{unique}$} and \bm{$f_{share}$} would serve as part of the input for the final classification layer as Eq. \ref{method:classification:9} in Sec. \ref{method:classification}. In this way, when the final classification objective is optimized, the image feature and text feature would be enforced to be projected into the same semantic space, and their cross-modal contrast would be assessed in this space by measuring the difference \bm{$T_u$} $-$ \bm{$I_u$}. In addition, the modal-shared content would also be fused with the knowledge information in the content-knowledge consistency subnetwork, which would be described in Sec~\ref{method:content-knowledge:fusion}.  
 


\subsection{Content-knowledge Consistency Subnetwork}\label{method:content-knowledge}

Here we introduce how to capture the content-knowledge inconsistency features.

\subsubsection{Entity pair sorting} 
After preprocessing in Sec. \ref{method:preprocessing}, the obtained entity representation is denoted as \bm{$e_l$} $\in \mathbb{R}^{d_e}$. We measure their Manhattan distance for each pair of entity representations within a post and retain the top-$k$ ($k=5$) entity pairs with the largest distances and their corresponding distance values. Note that for those posts where the number of entities is less than 4, the number of entity pairs can't reach 5 ($C_4^{2}=6$, $C_3^{2}=3$). To address this issue, we make a supplement with pseudo entities whose representations are random vectors. We concatenate the pairwise entity representations to get the entity pair representation $\boldsymbol{EP}_{i} \in \mathbb{R}^{2d_e}$ ($i \in [1,k]$). Also we get the  entity pair distance ${dis}^{i} \in \mathbb{R}$ ($i \in [1,k]$)


\subsubsection{Content-knowledge fusion with distance-ware signed attention} \label{method:content-knowledge:fusion}
To incorporate KG with post contents, we propose to fuse the top-$k$ largest-distance entity pairs with the modal-shared contents with the attention mechanism. We propose a novel approach that uses the modal-shared content as query $\boldsymbol{Q}$ and the entity pair representations $\boldsymbol{EP}$ as the value and key, and a distance-aware signed attention mechanism to learn the most relevant parts for fusion. By taking this approach, we can address the problem of content-knowledge consistency modeling and capture their complex semantic relationships.  This is different from the traditional usage of query, value and key in the attention mechanism as we can also capture the negative correlation between query and key. Moreover, unlike the originally signed attention in~\cite{tian2020qsan}, another factor (i.e., the entity distance) is taken into consideration to adjust the soft weights to better obtain content-knowledge inconsistency features.


We then illustrate the design of the distance-aware signed attention mechanism in detail. In the traditional attention mechanism, if the correlations between query and keys are negative (i.e., their compatibility (e.g., dot product) value is negative), we would treat it as insignificant. However, such a negative correlation may represent the opposing semantics that can be beneficial to the rumor detection task. Our signed attention mechanism, on the contrary, adds a ``-Softmax" operation using the opposite compatibility values between queries and keys as input to the Softmax function to amplify the negative correlations. Thus the compatibility values would go through two channels, that is, both the traditional Softmax (i.e., ``+Softmax'') and the ``-Softmax'' functions, to capture both positive and negative relationships between the modal-shared contents and the top-$k$ largest distance entity pairs. We thus obtain two attention weights corresponding to the two channels, that is, 


\begin{equation}
\begin{aligned}
    \boldsymbol{Q} &= {\rm Concat}(\boldsymbol{I_s}, \boldsymbol{T_s})\\
    \alpha_{pos}^i &= {\rm Softmax}\left(\frac{\boldsymbol{Q}\boldsymbol{(EP_{i})}}{\sqrt{2d_e}}\right)\\
    \alpha_{neg}^i &= -{\rm Softmax}\left(-\frac{\boldsymbol{Q}\boldsymbol{(EP_{i})}}{\sqrt{2d_e}}\right)
\end{aligned}
\end{equation}
where the modal-shared feature $\boldsymbol{Q}$ is the concatenation of modal-shared features for images and texts. 
Both $\alpha_{pos}^i$ and $\alpha_{neg}^i$ denote the attention weights of the $i$-th entity pair but reflect the positive and negative correlations, respectively. A larger $\alpha_{pos}^i$ (resp. $\alpha_{neg}^i$) means that the entity pair is more positively (resp. negatively) semantically related to the content. 

Meanwhile, an entity pair with a larger entity distance should influence the learning object more significantly. Following this intuition, we devise the final attention weight for each of the entity pairs by taking both of the factors into consideration and employ the weights to calculate the weighted sum of the entity pair representations, that is,



\begin{equation}
\begin{aligned}
     \beta_{*}^i &= \frac{dis^i\alpha_{*}^i}{\sum_{j = 1}^k dis^j*\alpha_{*}^j}\\
     \boldsymbol{f_{kg}^{*}} &= \sum_{i = 1}^k \beta_{*}^i (\boldsymbol{EP}_i)\\
     \boldsymbol{f_{kg}} &= {\rm Concat}(\boldsymbol{f_{kg}^{pos}},\boldsymbol{f_{kg}^{neg}}),
\end{aligned}    
\end{equation}
where $dis^i$ ($i \in [1,k]$) denotes the entity distance for the $i$-th entity pair, $\beta_{*}^i~(* \in \{pos,neg\})$ is the distance-aware signed attention weights, \bm{$f_{kg}^{*}$} $~(* \in \{pos,neg\})$ is the positive/negative entity-pair embedding based on the signed attention weights, an
\bm{$f_{kg}$} $\in \mathbb{R}^{4d_e}$ denotes the final semantic vector that represents the content-knowledge inconsistency features.




%


\subsection{Rumor Classification Layer}\label{method:classification}
Lastly, we concatenate the cross-modal inconsistency features, content-knowledge inconsistency features and the modal-shared features, and feed it into a fully-connected layer with Sigmoid activation function to obtain the predicted probability for instance $i$, that is,
\begin{equation}\label{method:classification:9}
    \hat{y}_i = \sigma(\boldsymbol{w_f}[\boldsymbol{f_{unique}}\oplus \boldsymbol{f_{share}}\oplus \boldsymbol{f_{kg}}] + \boldsymbol{b_f})
\end{equation}
where $\boldsymbol{w_f}$ and $\boldsymbol{b_f}$ are the weight and bias parameters. We then use cross-entropy loss as the rumor classification loss:
\begin{equation}
    \mathcal{L}_c = -\sum_{i} y_i log\hat{y}_i
\end{equation}
where $y_i$ is the ground truth label of the $i$-th instance. In addition, we also incorporate the orthogonal loss for multi-modal decomposition in Eq. \ref{method:cross-modal:5}. Thus, the final total loss is 
\begin{equation}
    \mathcal{L} = \mathcal{L}_c + \lambda\mathcal{L}_o
\end{equation}
where $\lambda$ is the weight of the orthogonal loss.



\section{Experiments}
\label{sec:length}
In this section, we conduct data analysis to validate the motivation that the dual-inconsistency information can be used to distinguish the rumors, and perform extensive experiments to evaluate the effectiveness of our proposal.

\subsection{Experimental Overview}

The experiments that we conduct can be divided into four parts: preliminary analysis, comparison experiments between our model and baselines, ablation studies, as well as robustness to different missing patterns. Since these experiments are conducted on either modal-incomplete or modal-complete datasets (or both of them), to make it clearer, we show which datasets correspond to which experiments in Table \ref{tab:index}.

For preliminary analysis, since we need to measure the cross-modal consistency to validate the statistically significant distinction between rumors and non-rumors, we conduct experiments on the modal-complete datasets. For comparison experiments, we perform experiments on both modal-incomplete and modal-complete datasets to validate that our
framework can outperform the baselines under both complete and incomplete modality conditions. Ablation studies are conducted on modal-incomplete datasets, since our model is mainly proposed for the real-world rumor detection scenario where visual modality is commonly missing. For the robustness experiments, we randomly mask some portion of the images, which is performed on the modal-complete datasets where the portion of images is gradually decremented from 100\% to 0\%.







\begin{table}
\caption{The correspondence between the datasets and the experiments.}
\label{tab:index}
\centering
\begin{tabular}{ccc}
\hline
\multirow{1}*{Expeiments}&\multicolumn{2}{c}{Datasets} \\ 
\cline{2-3}
&modal-incomplete & modal-complete\\
\hline
Preliminary analysis&&\checkmark    \\
Comparison experiments&\checkmark& \checkmark \\
Ablation studies&\checkmark&   \\
Robustness experiments &&\checkmark  \\
\hline
\end{tabular}
\end{table}


\subsection{Dataset}\label{experiments:dataset}
We conduct experiments on three real-world datasets, i.e., two English datasets: Twitter \cite{boididou2015verifying}, Pheme \cite{zubiaga2017exploiting} and one Chinese dataset: Weibo ~\cite{ma2016detecting}. Twitter and Pheme datasets are both collected from Twitter, while the Weibo dataset is collected from Weibo. The Twitter dataset is available at \url{https://github.com/MKLab-ITI/image-verification-corpus}. The Pheme dataset is available at \url{https://figshare.com/articles/PHEME_dataset_of_rumours_and_non-rumours/4010619}. The Weibo dataset is available at \url{https://www.dropbox.com/s/46r50ctrfa0ur1o/rumdect.zip?dl=0}
 As one primary objective of our proposal is to incorporate the post content and external knowledge information, we remove the data instances from which no entities can be extracted, as at least two entities are required in our model. As the statistics of the resulting datasets are shown in Table \ref{tab1:booktabs}, these three original datasets are all modal-incomplete. Note that if there are multiple images attached to one post, we randomly retain one image and discard the others. For the Twitter dataset, one image can be shared by various posts. 

To evaluate the performance of our model on the modal-complete dataset as well, we remove all the data instances from the original datasets without any images. We thus obtain three modal-complete datasets where both text and image are available for each post.
The statistics of the modal-complete datasets are also shown in Table \ref{tab1:booktabs}. It is obvious that these modal-complete datasets are subsets of the original modal-incomplete datasets.

\begin{table}
\normalsize
\caption{The statistics of the three original modal-incomplete datasets and three modal-complete datasets.}
\label{tab1:booktabs}
\centering
\resizebox{\linewidth}{!}{
\begin{tabular}{cc|cccccc}
\hline
\textbf{}& & \textbf{\#Posts} & \textbf{\#False} & \textbf{\#True} & \textbf{\#Posts w/ Image} &  \textbf{\#Entities/Post} \\
\hline
\multirow{1}*{Twitter}& modal-incomplete &18001 &11775 &6226 & 15557 & 5.302\\
&modal-complete & 15557 & 10184 & 5373 & 15557  & 5.536  \\
\hline
\multirow{1}*{Pheme} & modal-incomplete &5642 &1923 &3719 &2374 &4.383\\
& modal-complete & 2374 & 686 & 1688 & 2374  & 5.363 \\
\hline
\multirow{1}*{Weibo} &modal-incomplete &6691  &3542  & 3149 & 5450  & 3.232 \\
 &modal-complete& 5450 & 3104 & 2346 & 5450  & 3.557 \\
\hline
\end{tabular}
}

\end{table}

\subsection{Preliminary Analysis of Dual Inconsistency}\label{experiments:preliminary}
We conduct data analysis on the modal-complete datasets to validate that the two inconsistency metrics have statistically significant distinctions between rumors and non-rumors.

\begin{table}
\small
\caption{The average sum of the five largest entity distances and the average image-text similarity on three datasets. }\label{tab2:booktabs}
\centering
\resizebox{\linewidth}{!}{
\begin{tabular}{c|ccc|ccc}
\hline
 &\multicolumn{3}{c|}{Entity Distance}&\multicolumn{3}{c}{Image-text Similarity}\\
\cline{2-4}\cline{5-7}
 & Twitter &  Pheme& Weibo & Twitter & Pheme &Weibo  \\
\hline
 Rumors & 97.13 & 89.13 &99.98& -0.058 & -0.043&-0.063  \\
Non-rumors & 90.20  & 82.89  &96.31& 0.041  & 0.091&0.021  \\
\hline
\end{tabular}}

\end{table}

\subsubsection{Entity Distance Analysis}\label{experiments:preliminary:distance} 
We conduct entity distance analysis to show that the largest entity distances of a post are statistically different for rumors and non-rumors. Specifically, we measure the Manhattan distance of each pair of entity representations within a post and retain the top-$k$ ($k=5$) largest distance values (as described in Sec. \ref{method:content-knowledge}). The average sums of the five largest distances for all rumor and non-rumor posts are shown in Table \ref{tab2:booktabs}. We can observe that, on average, the sum of entity distances for rumors is larger than that for non-rumors. 

To statistically verify the observation, we make it a hypothesis and conduct hypothesis testing. For each dataset, two equal-sized collections of rumor and non-rumor tweets are sampled. And two-sample one-tail t-test is conducted on the 100 data instances to validate whether there is a sufficient statistical correlation to support the hypothesis. Let $\mu_f$ be the mean of the five largest entity distances of the rumor collection and $\mu_r$ represent that of non-rumors. The null hypothesis is $H_0$, and the alternative hypothesis is $H_1$. The hypothesis of interest is:
\begin{normalsize}
\begin{equation}
\begin{split}
    H_0&: \mu_f - \mu_r \leq 0 \\
    H_1&: \mu_f - \mu_r > 0
    \end{split}
\end{equation}
\end{normalsize}

The results show that there is statistical evidence
on all the datasets.
On Pheme, the result, t = $4.090$, df = $90$, p-value = $0.000047$ (significance alpha= $5\%$), rejects the $H_0$ hypothesis. The confidence interval CI is $[0.212, 42.112]$, the effect size is $0.826$. The conclusions are similar to Twitter and Weibo datasets.

\begin{table}
\caption{Comparison of different models from the perspective of modality used.}
\label{tab3_1:booktabs}
\centering
\begin{tabular}{cccc}
\hline
\multirow{1}*{Method}&\multicolumn{3}{c}{Modality} \\ 
\cline{2-4}
&Text&Image&KG\\
\hline
BERT&\checkmark& &    \\
Transformer&\checkmark& & \\
TextGCN&\checkmark& &  \\
EANN&\checkmark&\checkmark&  \\

SAFE&\checkmark&\checkmark&   \\
CompareNet&\checkmark&&\checkmark  \\
KMGCN&\checkmark&\checkmark&\checkmark \\
 \hline
KDCN Text-only &\checkmark&&\checkmark\\
\bf{KDCN}&\checkmark&\checkmark&\checkmark  \\
\hline
\end{tabular}
\end{table}

\subsubsection{Image-text Similarity Analysis}  
We also conduct the image-text similarity analysis towards rumors and non-rumors. In particular, we first decompose the raw textual and visual representations to obtain image-unique and text-unique embeddings excluding their shared information (refer to Eq. \ref{method:cross-modal:3} in Sec. \ref{method:cross-modal} for details) and measure their cosine similarity to get the image-text similarity. The average similarity results are shown in Table \ref{tab2:booktabs}. We can observe that the similarity for rumors is negative on all three datasets, while that for non-rumors is positive, so the similarity for rumors is much smaller than that for non-rumors, in line with our expectations. Moreover, we also perform hypothesis testing and confirm there is statistical evidence on all datasets. 

The rumor and non-rumor collections are set the same as Section \ref{experiments:preliminary:distance}. Let $\theta_f$ be the mean of cosine-similarity of the rumor collection and $\theta_r$ represents that of non-rumors. The null hypothesis is $H_0^{s}$, and the alternative hypothesis is $H_1^{s}$. The hypothesis of interest is:
\begin{normalsize}
\begin{equation}
\begin{split}
    H_0^{s}&: \theta_f - \theta_r \geq 0 \\
    H_1^{s}&: \theta_f - \theta_r < 0
    \end{split}
\end{equation}
\end{normalsize}

The results show that there are statistical evidence on the datasets. On Twitter dataset, the result, t = $-3.7925$, df = $97$, p-value = $0.000129$ ( significance alpha= $5\%$), rejects the $H_0$ hypothesis. The confidence interval CI is $[-0.425888, -0.002151]$, the effect size is $-0.7662$. We also found statistical evidences on Pheme dataset, with t = $-7.9051$, df = $94$, p-value = $2.4769\times 10^{-12}$ ( significance alpha= $5\%$), rejects the $H_0$ hypothesis. The confidence interval CI is $[-0.317446, -0.001603]$, the effect size is $-1.5970$. On the Weibo dataset, the results are t = $-2.8743$, df = $93$, p-value = $0.0025$ (significance alpha= $5\%$), rejects the $H_0$ hypothesis. The confidence interval CI is $[-0.001603,-0.317373]$, the effect size is $-0.5807$.
Our analysis shows that on each dataset, the rumors own distinct content-knowledge inconsistency and cross-modal inconsistency from non-rumors, which can help distinguish rumors from non-rumors. 

In the above data analysis as well as the methodology section, we consider top-$k$ ($k=5$) largest distances between entities, rather than averaging distances between all entity pairs, as the latter would weaken the contrast between rumors and non-rumors. The gap between the average distances of non-rumors and rumors would decrease significantly by the increase of $k$ in preliminary analysis. When $k>5$, 
the average distances between non-rumors and rumors become marginal.
This is because even for rumors, there are still some consistent entities. For the example in Fig. \ref{fig1_example}, a shark that appears in water is reasonable, and a subway station usually has elevators. In addition, since some posts have few entities, a larger $k$ may lead to the adoption of more pseudo entities in our framework, which may introduce larger noises. We later empirically show in Fig. \ref{fig_ablation} that considering top-5 can achieve good performance.

\begin{table*}
\caption{Results of comparison among different models on Pheme, Weibo and Twitter Datasets under modal-incomplete and modal-complete conditions.
The best performance per dataset is shown in {\bf bold}, while the runner-up performance is underlined.}
\label{tab3:booktabs}
\centering
\resizebox{\textwidth}{!}{
\begin{tabular}{cccccccccccc}
\toprule
\multicolumn{2}{l}{Datasets}&Metric &Bert & Transformer & TextGCN & EANN & SAFE & CompareNet & KMGCN & \makecell[c]{KDCN \\ Text-only} & KDCN  \\
\hline
&& Acc. & 0.817 & 0.789 & 0.826 & 0.815 & 0.786 & 0.750 & 0.825 & \underline{0.848} & \bf{0.849} \\
&& Prec. & 0.816 & 0.773 & 0.806 & 0.799 & 0.775 & 0.750 & 0.806 & \underline{0.833}& \bf{0.836}\\
&modal-incomplete & Rec. & 0.764 & 0.799 & 0.821 & 0.771 &0.554  & 0.750 &0.804  &\bf{0.837} &\underline{0.827}\\
&& F1. & 0.789 & 0.785 & 0.813 & 0.782 & 0.646 &0.750  & 0.805 &\bf{0.835} &\underline{0.831}\\
\cline{2-12}
Pheme&& Acc. &0.819  & 0.774 & 0.810 & 0.766 &0.782  & 0.765 & 0.812 & \underline{0.842} & \bf{0.862} \\
&& Prec. & 0.809 & 0.755 & 0.775 & 0.701 & 0.635 & 0.765 & 0.775 & \underline{0.811} & \bf{0.833}\\
&modal-complete & Rec. & 0.726 & 0.648 & 0.744 & 0.687 & 0.515 & 0.765 & 0.753 &\underline{0.802} &\bf{0.831}\\
&& F1. & 0.765 & 0.697 & 0.759 & 0.693 & 0.569 & 0.765 & 0.764 &\underline{0.806} &\bf{0.832}\\
\hline\hline
& & Acc. & 0.912 & 0.832 & 0.878 & 0.836 &0.906  &0.850  &0.881 & \underline{0.919} & \bf{0.924} \\
&& Prec. & 0.912 & 0.832 & 0.878 & 0.837 & 0.902 & 0.850 & 0.881 & \underline{0.919}& \bf{0.924}\\
&modal-incomplete & Rec. & 0.913 & 0.831 & 0.878 & 0.836 & 0.906 & 0.850 & 0.880 &\underline{0.919} &\bf{0.923}\\
&& F1. & 0.913 & 0.831 & 0.878 & 0.836 & 0.904 &0.850  & 0.880 &\underline{0.919} &\bf{0.924}\\
\cline{2-12}
Weibo& & Acc. & 0.881  & 0.772 & 0.860 & 0.788 &0.895  & 0.833 & 0.861 &\underline{0.925}  & \bf{0.943} \\
&& Prec. & 0.886 & 0.779 & 0.871 & 0.786 &0.915  &0.833  & 0.864 &\underline{0.925} & \bf{0.941}\\
&modal-complete & Rec. & 0.881 & 0.772 & 0.861 & 0.791 & 0.897 & 0.833 & 0.856 &\underline{0.925} &\bf{0.943}\\
&& F1. & 0.884 & 0.775 & 0.866 & 0.786 & 0.906 &0.833  & 0.860 &\underline{0.925} &\bf{0.942}\\
\hline\hline
 && Acc. & 0.892 & 0.822 & 0.839 & 0.796 & 0.867 &0.826  &0.846  & \underline{0.901} & \bf{0.931} \\
&& Prec.& \underline{0.894} & 0.803 & 0.823 & 0.729 &0.876  &0.825  &0.829  &0.890 & \bf{0.917}\\
&modal-incomplete & Rec. & 0.863 & 0.819 & 0.849 & 0.719 &\underline{0.927} & 0.782 & 0.852 &0.892 &\bf{0.941}\\
&& F1. & 0.879 & 0.811 & 0.836 &0.724  & \underline{0.901} &0.796  & 0.840 &0.891 &\bf{0.929}\\
\cline{2-12}
Twitter&& Acc. & 0.835 & 0.791 & 0.712 & 0.697 &\underline{0.843}  &0.823  & 0.825 & 0.837 & \bf{0.945} \\
&& Prec.& 0.821 & 0.772 & 0.721 & 0.695 & \underline{0.847} &0.823  & 0.813 & 0.796& \bf{0.946}\\
&modal-complete & Rec. & 0.810 & 0.791 & 0.744 & 0.698 & \underline{0.851} & 0.783 & 0.788 &0.814 &\bf{0.916}\\
&& F1. & 0.815 & 0.781 & 0.732 & 0.697 &\underline{0.849}  &0.796  & 0.800 &0.805 &\bf{0.931}\\

\bottomrule
\end{tabular}}
\end{table*}

\subsection{Experimental Setup}\label{experiments:setup}

In all experiments, we randomly split the Pheme and Weibo datasets 
into training, validation, and testing sets with a split ratio of 6:2:2 without overlapping, and conduct a 5-fold cross-validation to obtain the final results. For the Twitter dataset, since it has an official data splitting when publishing, we follow its splitting ratio (approximately 8:1:1) and don't apply 5-fold cross-validation. All the data splittings have ensured that images in the training set and testing set will not be overlapped.

Our algorithms are implemented on Pytorch framework~\cite{paszke2017automatic} and trained with Adam~\cite{kingma2014adam}. In terms of parameter settings, the learning rate is \{0.0005, 0.00005\}, and batch size is \{64, 128\}. 
The weight of the orthogonal loss is $\lambda=1.5$.  
We adopt an early stop strategy and dynamic learning rate reduction for model training.

We use the pre-trained BERT~\cite{wolf-etal-2020-transformers} as initial word embeddings for text encoding in our model: bert-base-uncased for English, and bert-base-chinese for Chinese. For other models that don't adopt BERT, we use GloVe~\footnote{GloVe: Global Vectors for Word Representation:\url{https://nlp.stanford.edu/projects/glove/}} instead.

\subsection{Baselines}\label{experiments:baselines}
The baselines are listed as follows:\\
\begin{itemize}
\item{\bf BERT} ~\cite{devlin2018bert}  is a pre-trained language model based on deep bidirectional transformers, and we use it to get the representation of the post text for classification. We use BERT with fine-tuning to detect rumors, which is available at \url{https://github.com/huggingface/transformers}.

\item{\bf Transformer} ~\cite{vaswani2017attention} uses the self-attention mechanism and position encoding to extract textual features for sequence to sequence learning. We only use its encoder here. we use the publicly available implementation at \url{https://github.com/jayparks/transformer}.
\item{\bf TextGCN}~\cite{yao2019graph} uses a graph convolution network to classify documents. The whole corpus is modeled as a heterogeneous graph to learn the word and document embeddings. 
The heterogeneous graph contains word nodes and document nodes. The edges are built based on word occurrence and document word relations. We use the publicly available implementation at \url{https://github.com/chengsen/PyTorch_TextGCN}.

\item{\bf EANN}~\cite{wang2018eann} uses an event adversarial neural network to extract event-invariant features from images and texts for rumor detection. For modal-incomplete instances, we use white images to supplement. We used the authors' implementation, which is available at \url{https://github.com/yaqingwang/EANN-KDD18}.  
\item {\bf SAFE} ~\cite{zhou2020mathsf} is a similarity-aware fake news detection method. It extracts textual and visual features for news and then further investigates the relationship between the extracted features across modalities. For modal-incomplete instances, we use white images to supplement. We used the authors' implementation, which is available at \url{https://github.com/Jindi0/SAFE}.
\item {\bf CompareNet} ~\cite{hu-etal-2021-compare} proposes a graph neural model, which compares the news to the knowledge base (KB) through entities for fake news detection. We used the authors' implementation, which is available at \url{https://github.com/BUPT-GAMMA/CompareNet_FakeNewsDetection}.
\item{\bf KMGCN}~\cite{wang2020fake} is a state-of-the-art rumor detection model that uses a graph convolution network to incorporate visual information and KG to enhance the semantic representation. Since the authors don't release the code, we implemented the method. We followed the implementation details described in KMGCN except for choosing a different KG. Instead of using Probase and Yago in the original KMGCN, we used Freebase as the reference knowledge graph and acquired isA relation of the entities, to make a fair comparison with our model. The Freebase isA relation data dump is available at \url{https://freebase-easy.cs.uni-freiburg.de/dump/}
\item {\bf KDCN Text-only} is our full model but trained using the single-modal text data only, replacing all the input images with white images. It represents an extremely modal-incomplete condition that all the images are missing. 

\end{itemize}

Table \ref{tab3_1:booktabs} compares the baselines and the proposed model KDCN from the perspective
of the modality data that are used. All baseline models and our model can be grouped into four categories: models using only text modality, models using both text and image data, models using text and knowledge data, and models using text, image, and knowledge data. Note that since EANN and SAFE require images as input and cannot adapt to modal-missing conditions, we also use white images as supplementary in modal-incomplete cases, which is the same as our model for a fair comparison.  

\begin{figure*}
    \centering
    \subfigure[\emph{Pheme} dataset]{\includegraphics[width=0.33\linewidth]{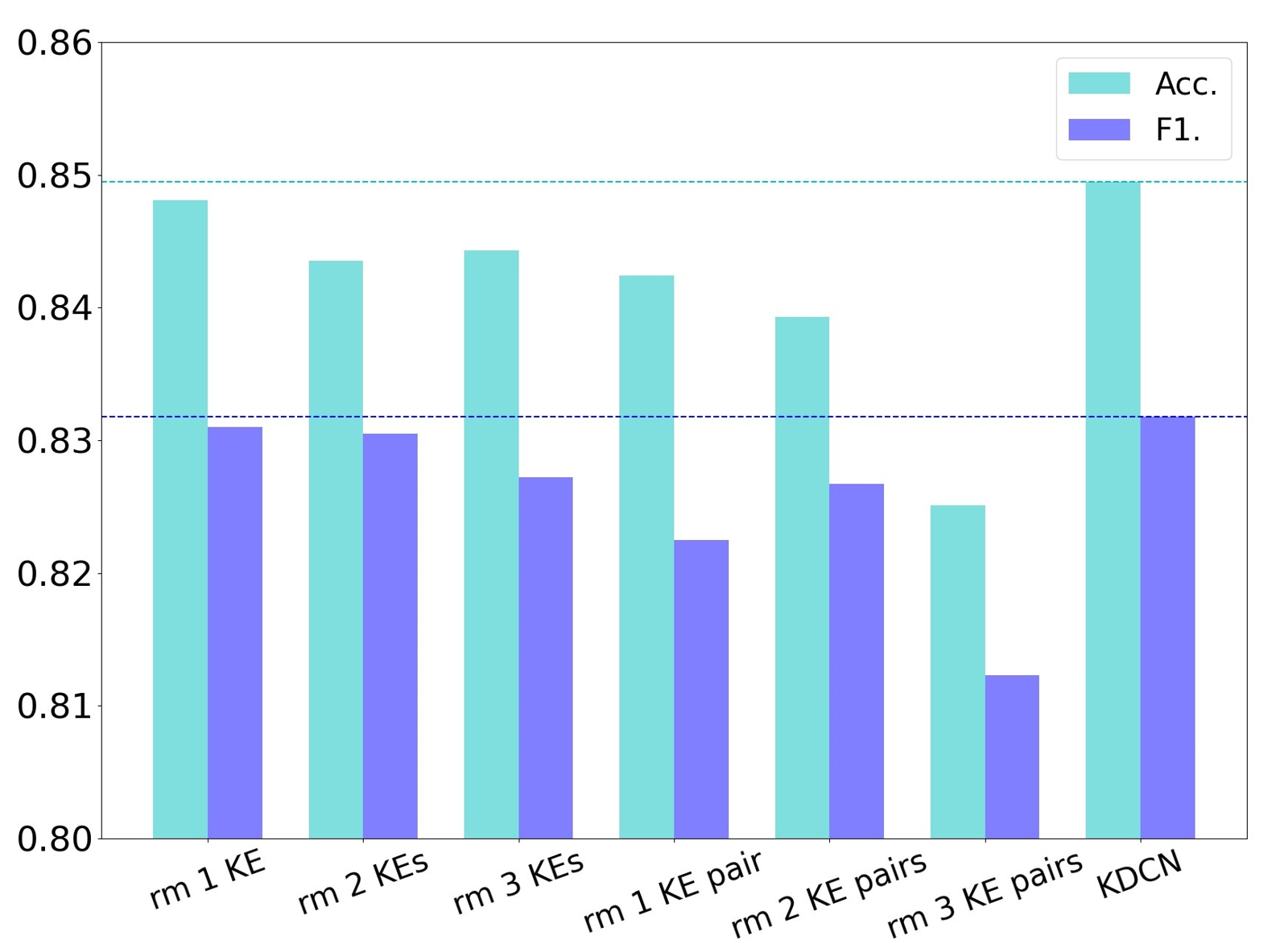}}
    \subfigure[\emph{Weibo} dataset]{\includegraphics[width=0.33\linewidth]{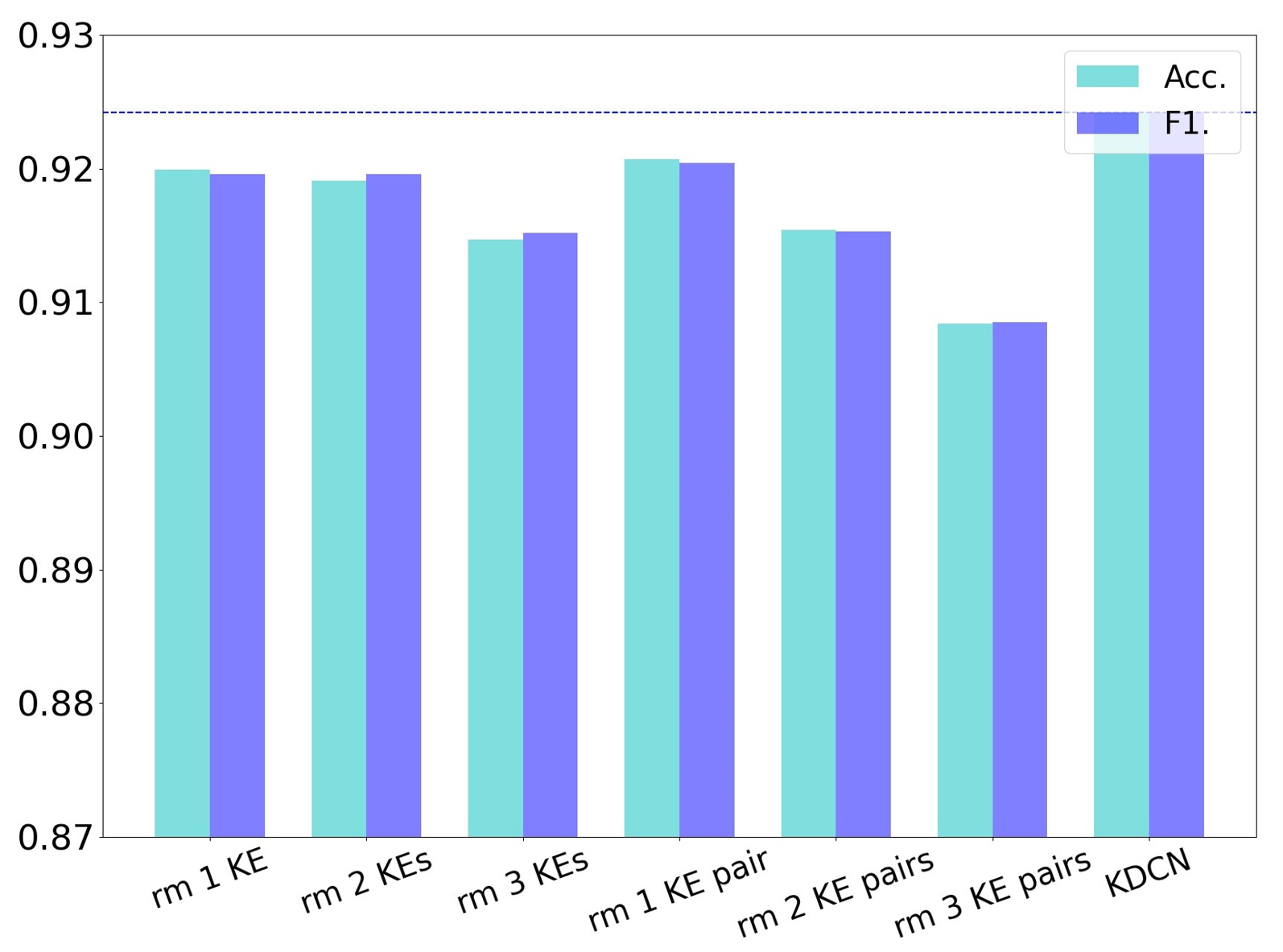}}
    \subfigure[\emph{Twitter} dataset]{\includegraphics[width=0.33\linewidth]{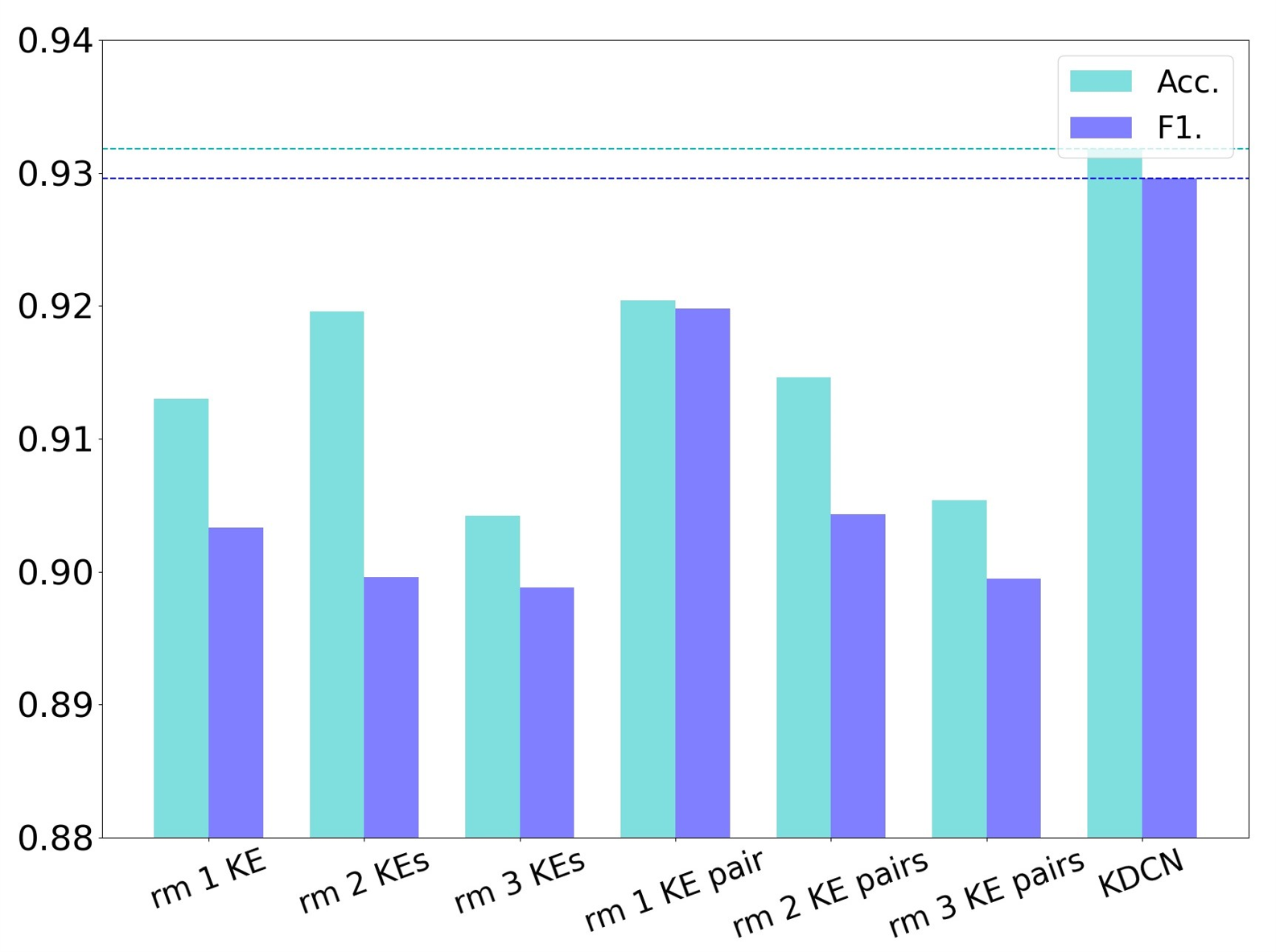}}
    \caption{Results of the sensitivity analysis with varying number of entities and entity pairs on Pheme, Weibo and Twitter datasets under the modal-incomplete condition. The two horizontal lines indicate accuracy and F1 values of the proposed model KDCN.}
    \label{fig_ablation}
\end{figure*}

\subsection{Results and Discussion}\label{experiments:results}
Table \ref{tab3:booktabs} demonstrates the performance of all the compared models on three datasets. We can observe that under both modal-incomplete and modal-complete conditions, our model \textbf{KDCN} generally significantly outperforms all the baselines in all the metrics, which confirms that considering the two inconsistencies would benefit the rumor detection task.

Among the three state-of-the-art textual representation models, BERT outperforms both Transformer and TextGCN on Weibo and Twitter datasets under modal-incomplete conditions. While under the modal-complete condition, BERT outperforms the other two on all three datasets, demonstrating its superior capability in capturing the textual semantics for rumor detection.

We then compare the models involving the visual information with the above text-only models. Although EANN considers both visual and textual information, it performs not as well as BERT and TextGCN under both modal-incomplete and modal-complete conditions. The possible reason is that EANN uses CNN to extract the textual feature, which is not as powerful as Transformer and GCN. SAFE outperforms EANN in most cases, indicating that the text-image dissimilarity captured in SAFE is an effective feature for rumor detection. 


 KMGCN achieves comparable or better performance compared to TextGCN and CompareNet under both modal-incomplete and modal-complete conditions. Since all these three models adopt graph convolution networks as the backbone, it indicates that the image and knowledge features can provide complementary information and improve performance. 

Despite the lack of visual information, KDCN Text-only performs better than textual representation models, and achieves the runner-up performance in most cases, indicating that the content-knowledge inconsistency can enhance the model performance.

Compared to the baselines, we can attribute our proposal's superiority to three critical properties: (1) we model two types of inconsistent information, which are suitable to rumor identification; (2) we adopt BERT as the initial text representation to capture textual semantics; (3) we adopt the complete-modality token to make the model applicable for visual modality missing conditions and achieve robust performance. 

Please note that to address the visual-modality missing issue, we also have tried to generate images based on the corresponding text content using generative adversarial networks, and it achieves comparable performance as using the white image with a special [CMT] token. In particular, its performance on the Pheme-incomplete dataset is 0.8438 and 0.8382 in terms of Acc. and F1, respectively. Despite the similar performance as our proposal, using generative adversarial networks would incur heavy computational costs. We also have tried to use randomly generated images as a complement, and the performance on the Pheme-incomplete dataset is 0.8099 in terms of Acc., which is much lower than our proposal. The possible reason is that it introduces noises that are entirely unrelated to the text.

\begin{figure*}
    \centering
    \subfigure[\emph{Pheme} dataset]{\includegraphics[width=0.33\linewidth]{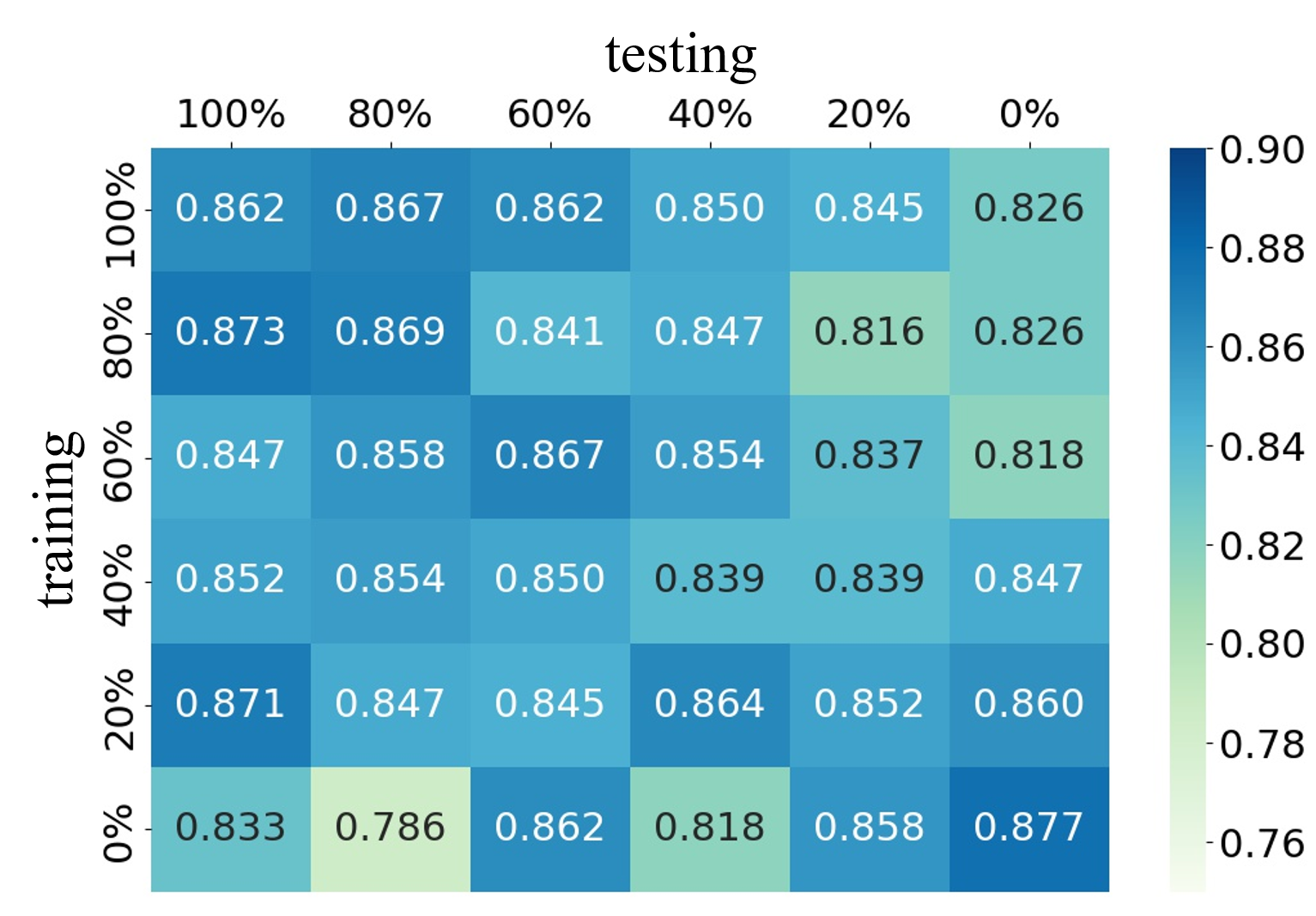}}
    \subfigure[\emph{Weibo} dataset]{\includegraphics[width=0.33\linewidth]{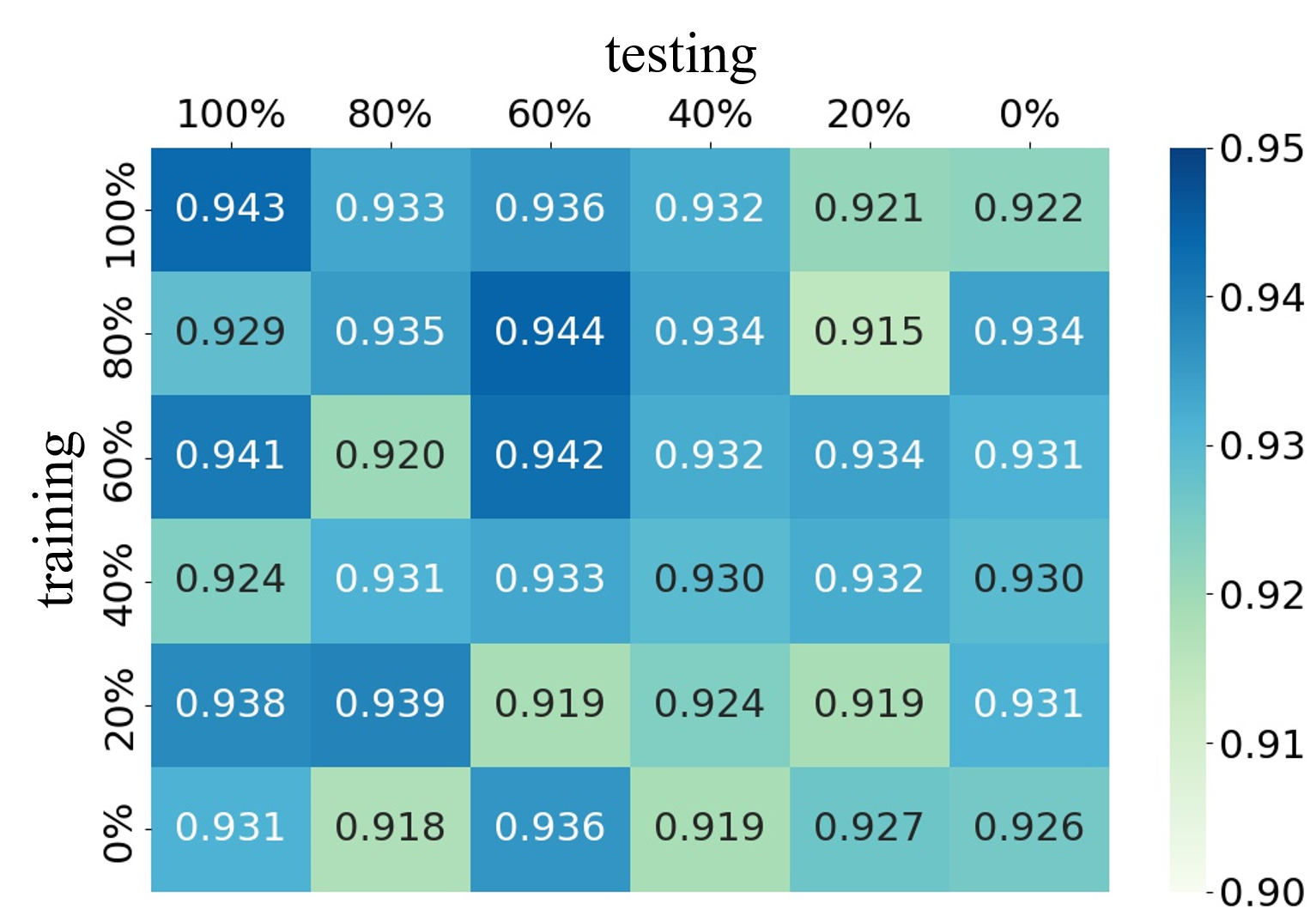}}
    \subfigure[\emph{Twitter} dataset]{\includegraphics[width=0.33\linewidth]{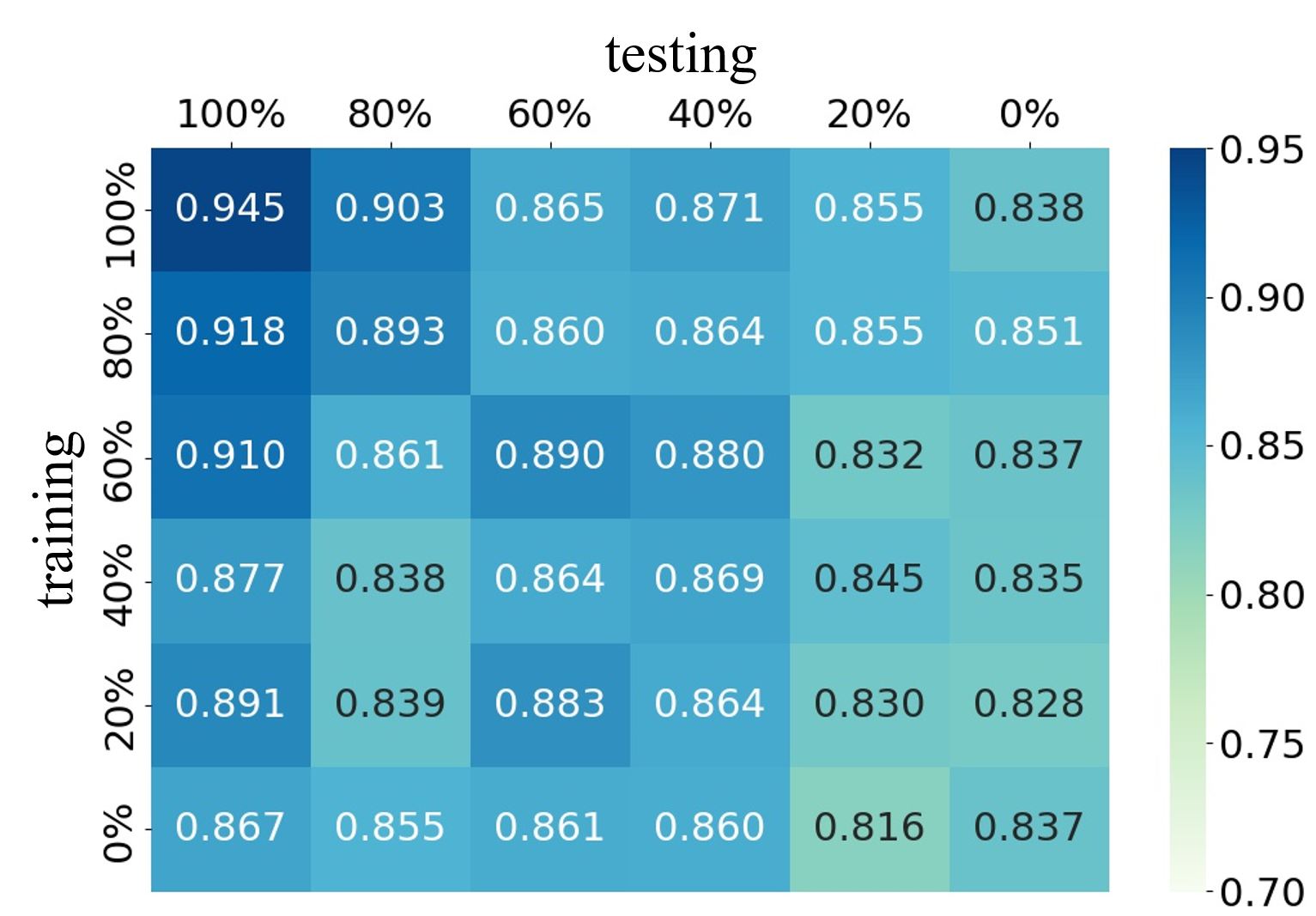}}
    \caption{Classification accuracy on Pheme, Weibo and Twitter datasets with different missing patterns. The row (resp. column) of the matrix represents the percentage of the training (resp. testing) instances that are equipped with the visual data. The darker the blue, the higher the accuracy.}
    \label{fig_missing}
\end{figure*}

\begin{figure}[htbp]
    \centering
    \subfigure[Zombie apocalypse approaches RT @thinkprogress: Sandy approaches NYC Sandy hurricane.]{\includegraphics[height=2.5cm,width=0.4\linewidth]{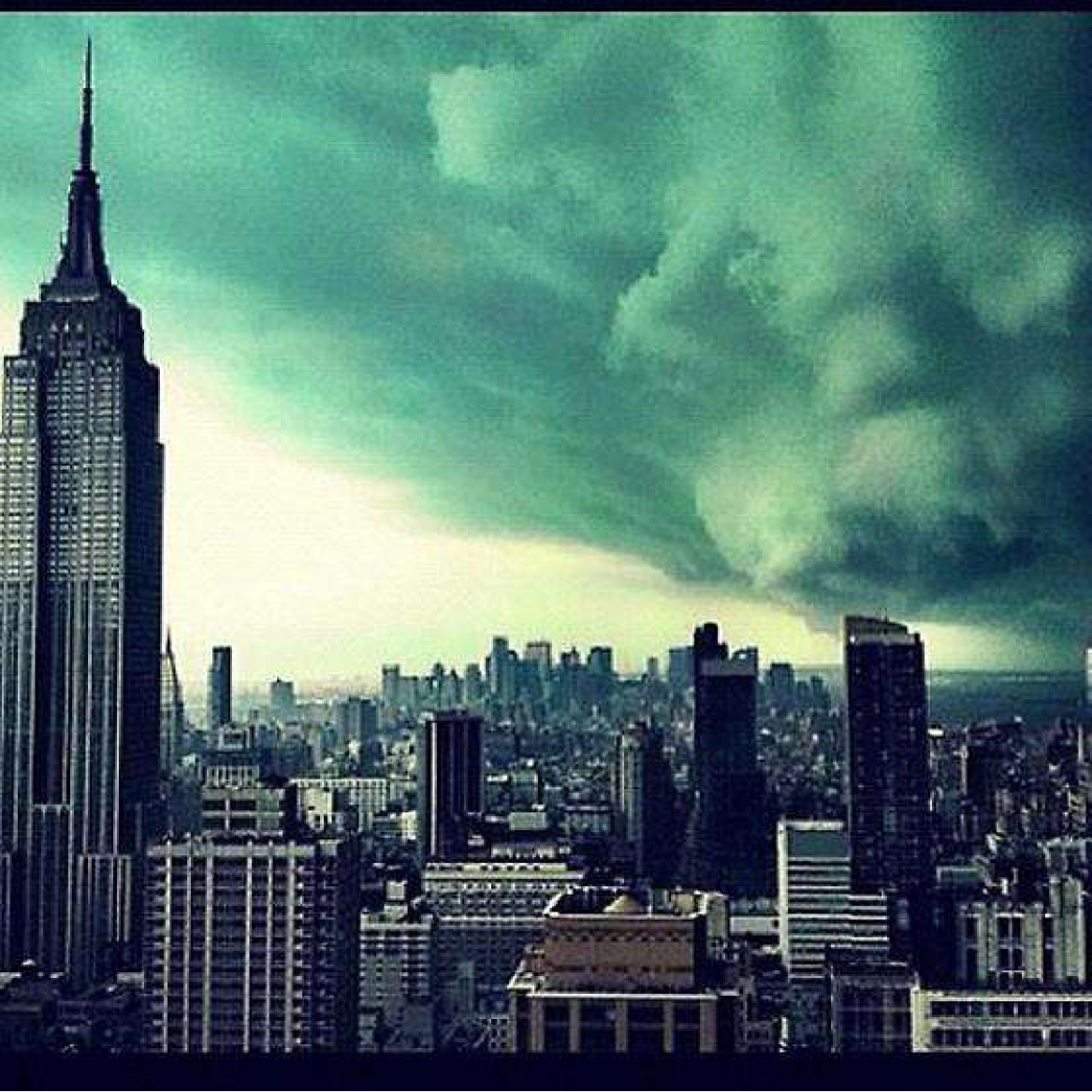}}\hspace{7mm}
    \subfigure[NHL postpones Maple Leafs-Senators game after tragic shootings in Ottawa.]{\includegraphics[height=2.5cm,width=0.4\linewidth]{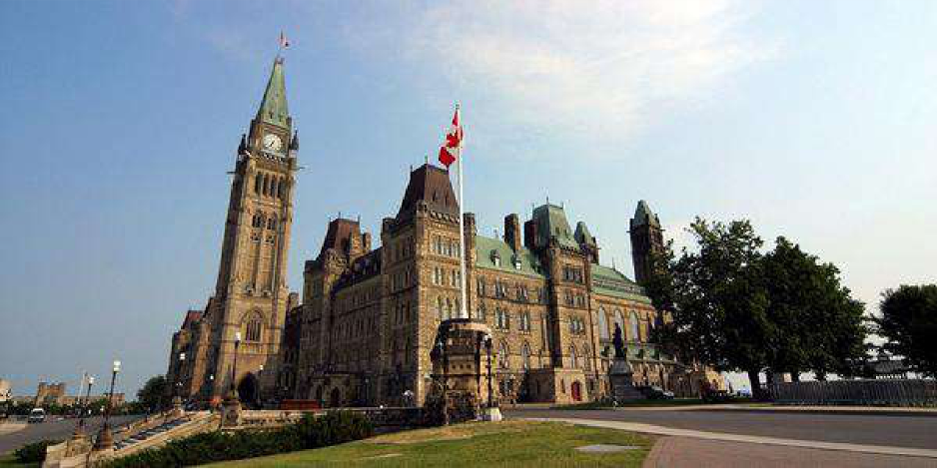}}
    \caption{Two rumor cases detected by our model.}
    \label{fig:rmy_label}
\end{figure}

\begin{table}
\centering
\normalsize
\resizebox{\linewidth}{!}{
\begin{tabular}{lcccccc}
\hline
\multirow{1}*{Method} & \multicolumn{2}{c}{Pheme} &  \multicolumn{2}{c}{Weibo} &  \multicolumn{2}{c}{Twitter}  \\
\cline{2-3}\cline{4-5}\cline{6-7}
& Acc. & F1. & Acc. & F1.  & Acc. & F1.\\
\hline
KDCN & {\bf 0.849} & {0.831}  & {\bf 0.924} & {\bf 0.924} & {\bf 0.931}&{\bf 0.929} \\
\hline
\ -w/o Visual & 0.846 & {\bf 0.836}  & 0.918 & 0.918 &0.907 & 0.902\\
\ -concat. TV & 0.836 & 0.821 & 0.922 & 0.922 & 0.917&0.912 \\
\ -w/o KE & 0.832 & 0.817  & 0.921 & 0.921 & 0.908&0.898 \\
\ -mean KE & 0.843 & 0.826 & 0.921 & 0.922&0.930 &0.925\\
\ -w/o CMT &0.844 &0.829 & 0.922&0.923 &0.921 &0.912\\
\ -w/o Orthog. Loss & 0.839&0.823 &0.919 & 0.920& 0.923&0.920 \\
\hline
\end{tabular}}
\caption{Results of comparison among different variants on modal-incomplete Pheme, Weibo and Twitter datasets.}
\label{tab4:booktabs}
\end{table}
\subsection{Performance of the Variations}\label{experiments:ablation}
We investigate the effects of our proposed components by defining the following variations:
\begin{itemize}

\item{\bf w/o Visual}: the variant that removes the visual information.
\item {\bf concat. TV:} the variant that concatenates the textual and visual representations instead of their cross-modal inconsistency and modal-shared features.
\item {\bf w/o KE}: the variant that removes the content-knowledge consistency subnetwork.
\item {\bf mean KE:} the variant that utilizes the mean pooling of the entity representations instead of the content-knowledge inconsistency features.
\item {\bf w/o CMT}: the variant that removes the complete-modality token ([CMT]). Then Equation (2) would be $ H_I = \mathbf{ReLU}(\boldsymbol{w_I}*(\mathbf{CNN}(Image))+\boldsymbol{b_I})$.
\item {\bf w/o Orthog. Loss}:  the variant that removes the orthogonal loss from the final total loss, with only the cross entropy loss left.

\end{itemize}
 The ablation study in Table \ref{tab4:booktabs} demonstrates that the proposed components are indispensable for achieving the best performance. Visual features can improve performance. To further show the effectiveness of the inconsistency features, we use the same input but alternate aggregating mechanisms, i.e., \emph{mean KE} and \emph{concat. TV}, instead of the proposed inconsistency mechanisms. We can observe that the results of both \emph{mean KE} and \emph{concat. TV} are lower than the proposed model, indicating that the inconsistency features are more effective than the aggregated features for rumor detection. 
\emph{w/o Orthog. Loss} also yields worse performance than the proposed model, suggesting that the constraint on the decomposed modal-unique and modal-share spaces is beneficial for the model to learn a better representation of multi-modal data. The results of \emph{w/o CMT} are lower than the KDCN model, indicating that the addition of the [CMT] token does help the model distinguish between the presence and absence of the visual modality.

 
To verify the effectiveness of the knowledge information, we conduct the sensitivity analysis with a varying number of entities and entity pairs, and design the following variants:
\begin{itemize}
	\item{\bf rm $n$ KE:} the variant that randomly removes $n$ ($n \in \{1,2,3\}$) entities from the post entity set.
	\item{\bf rm $n$ KE pair:} the variant that randomly removes top-$n$ ($n \in \{1,2,3\}$) largest distance entity pairs from the post entity set.
\end{itemize}
As shown in Fig. \ref{fig_ablation}, it can be observed that the accuracy decreases gradually as more entity pairs are removed in the content-knowledge consistency subnetwork. Similar trends can be observed when one or more entities are removed. It verifies the crucial impact of the knowledge information for our task. 

It can be observed that the performance degradation when removing the entities and entity pairs on the Weibo dataset is not as large as on the other two datasets. The possible reason is that the number of extracted Chinese entities is not as large as the other two English datasets due to the limited coverage of KG on Chinese entities. In particular, as shown in Table \ref{tab:index}, the column of ``Entities/Post” shows the average number of entities in one post for these datasets, and we can see that Weibo has the lowest number. In fact, for Weibo-incomplete and Weibo-complete datasets, the average number of entities in one post is nearly 3. Since we measure the Manhattan distance for each pair of entity representations within a post and retain the top-5 entity pairs with the largest distances, for the above cases when the number of entity pairs cannot reach 5 ($C_4^2$ = 6, $C_3^2$ = 3), we would make a supplement with pseudo entities whose representations are random vectors. It may introduce noises and cannot achieve better performance. This suggests that we can utilize a larger-scale KG and more powerful entity-extracting techniques to further improve performance in future work.

\subsection{Robustness to Different Missing Patterns }\label{experiments:robust}
To verify the robustness of our model against the visual modality missing issue, we conduct experiments under different missing patterns.


\textbf{Setting of different missing patterns.}  We randomly mask some portion of the images in the modal-complete datasets (Twitter-mc, Pheme-mc and weibo-mc) to produce different visual-modality missing datasets. 
Specifically, we produce the following missing patterns:
training with 100\% Text + $\eta$ \% Image and testing with 100\% Text + $\mu$\% Image. $\eta$ and $\mu$ $\in$ [0,20,40,60,80,100].


\textbf{Results of Robustness to Different Missing Patterns.} Fig. \ref{fig_missing} shows the results of our approach under the different missing patterns. We have two observations. Firstly, the rumor detection performance of our model is quite stable under different missing patterns. Moreover, despite the lack of visual data, most of these results are still better than the baselines with full-modal data as shown in Table \ref{tab3:booktabs}. Secondly, according to Fig. \ref{fig_missing}, as the $\eta$ and $\mu$ are larger, the blue color of the entry generally becomes darker. It indicates that our model would perform better when more visual data is available.

\subsection{Case Studies}\label{experiments:case}

We analyze two rumor cases that our model can recognize accurately. They are from Twitter and Pheme, respectively. In Fig~\ref{fig:rmy_label} (a), the extracted entity set is \emph{\{Zombie, Tropical cyclone, New York City, RT (TV network), ThinkProgress\}}. The average sum of the five largest entity distances is 119.73, larger than the average sum of the rumors on Twitter (i.e., 97.13 shown in Table \ref{tab2:booktabs}), implying the existence of content-knowledge inconsistency. Its image-text similarity value is 0.277, much larger than the average value for rumors (-0.058 in Table \ref{tab2:booktabs}), indicating the image and text are well matched. In Fig~\ref{fig:rmy_label} (b), it is obvious that the image and text are not well-matched, verified by its low image-text similarity value (only -0.133). The two cases help to confirm that our model can effectively capture the two types of inconsistent information for rumor identification.

\section{Conclusion}

We propose a knowledge-guided dual-consistency network for multi-modal rumor detection, which involves the cross-modal inconsistency and content-knowledge inconsistency information in one framework. Additionally, our framework can also deal with visual modality issues in real-world detection scenarios. Extensive experiments on three datasets have demonstrated our proposal's effectiveness in capturing and fusing both types of inconsistent features to achieve the best performance, under both modal-complete and modal-incomplete conditions. Note that the inconsistent features captured by our framework can be easily plugged into other rumor detection frameworks to further improve their performance. In future work, we plan to explore more effective inconsistency features and devise a more explainable and robust model.



%



\ifCLASSOPTIONcompsoc
  \section*{Acknowledgments}
\else
  \section*{Acknowledgment}
\fi
This work was supported by the Natural Science Foundation of China (No.61976026) and 111 Project (B18008). This work was supported in part by NSF under grants III-1763325, III-1909323,  III-2106758, and SaTC-1930941. Sihong was supported in part by the National Science Foundation under NSF Grants IIS-1909879, CNS-1931042, IIS-2008155, and IIS-2145922. Any opinions, findings, conclusions, or recommendations expressed in this document are those of the author(s) and should not be interpreted as the views of any U.S. Government. The entity linking solution TAGME exploited in this work has been supported by services offered via the D4Science Gateway (\url{https://services.d4science.org/}) operated by D4Science.org (\url{www.d4science.org}).

\ifCLASSOPTIONcaptionsoff
  \newpage
\fi



%


\bibliographystyle{IEEEtran}
\bibliography{tkde-ref}

%

\begin{IEEEbiography}[{\includegraphics[width=1in,height=1in,clip]{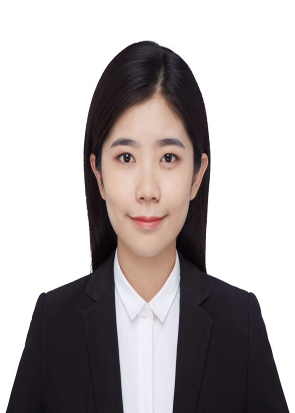}}]{Mengzhu Sun}
received the bachelor's degree in Mathematics and Applied Mathematics from Beijing Normal University, in 2019. She is working toward the master's degree in the Key Laboratory of Trustworthy Distributed Computing and Services, Beijing University of Posts and Telecommunications, Ministry of Education, China. Her research interests include data mining and machine learning.
\end{IEEEbiography}

\begin{IEEEbiography}[{\includegraphics[width=1in,height=1.25in,clip,keepaspectratio]{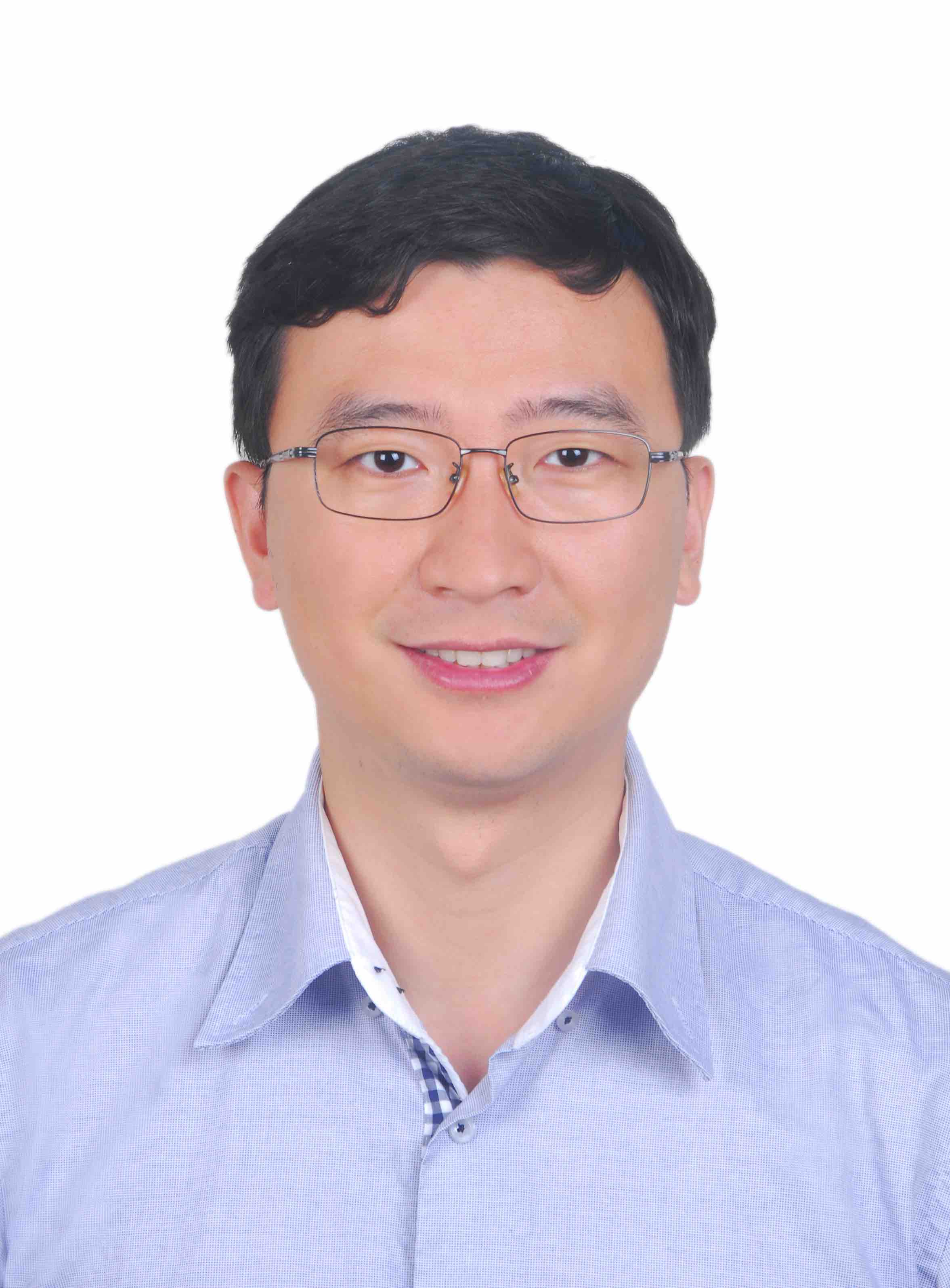}}]{Xi Zhang}
received the Ph.D. degree in computer science from Tsinghua University. He is a professor at the Beijing University of Posts and Telecommunications, and is also the vice director of the Key Laboratory of Trustworthy Distributed Computing and Service, Ministry of Education, China. He was a visiting scholar at the University of Illinois at Chicago. His research interests include data mining and computer architecture. He is a member of the IEEE.
\end{IEEEbiography}

\begin{IEEEbiography}[{\includegraphics[width=1in,height=1.25in,clip,keepaspectratio]{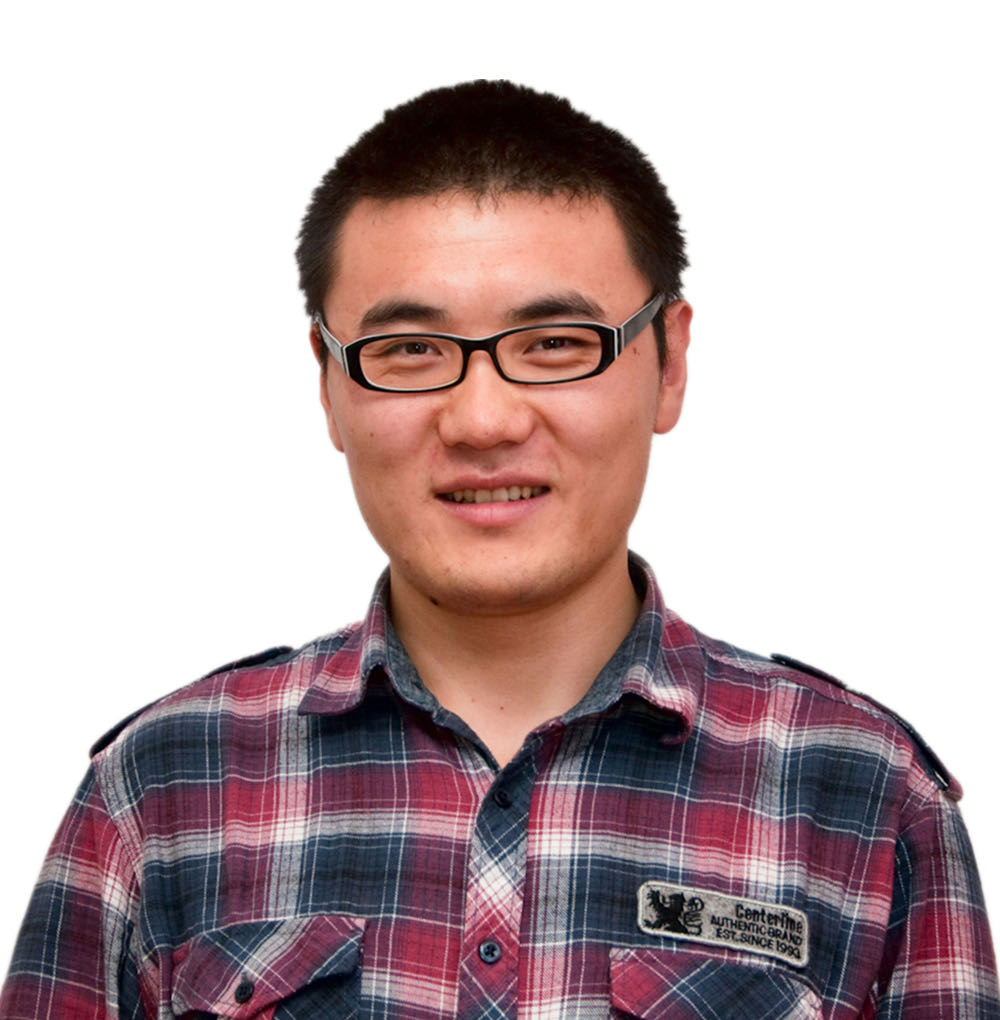}}]{Jianqiang Ma} is currently a staff researcher at Tencent Video, where he leads the knowledge graph application and retrieval algorithm development of the search product. His main research interests include natural language processing, knowledge graph and IR. Before joining Tencent, he worked at Ping An Group, an AI startup and the University of Tübingen, where he was a Marie Curie fellow. He received the BE in computer science from Harbin Institute of Technology and the MA in Language and Communication Technologies from the University of Groningen
\end{IEEEbiography}

\begin{IEEEbiography}[{\includegraphics[width=1in,height=1.25in,clip,keepaspectratio]{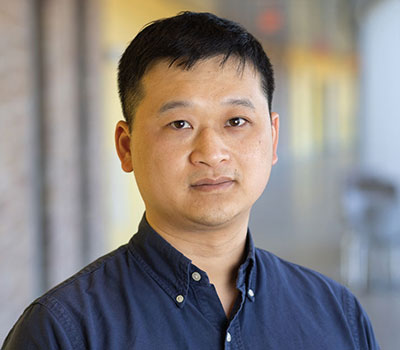}}]{Sihong Xie} is an assistant professor at the Department of Computer Science and Engineering, Lehigh University, Pennsylvania, U.S.
He received his Ph.D. in 2016 from the Department of Computer Science at the University of Illinois at Chicago, under the supervision of Philip S. Yu.
His research interest includes accountable graphical models, misinformation detection in adversarial environments, and human-ML collaboration in structural data annotation.
He received an NSF CAREER award in 2022.
\end{IEEEbiography}

\begin{IEEEbiography}[{\includegraphics[width=1in,height=1.25in,clip,keepaspectratio]{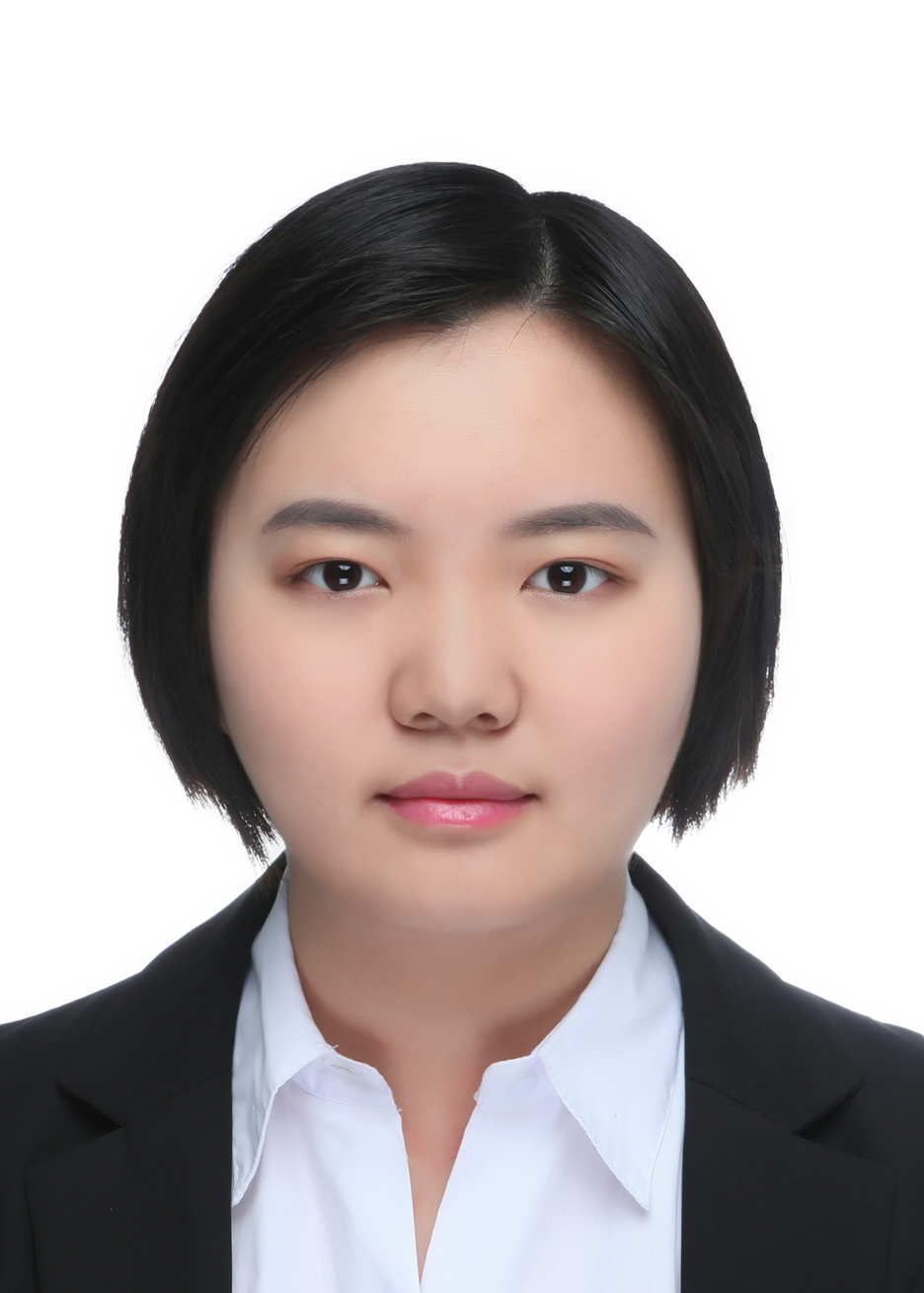}}]{Yazheng Liu}
received the bachelor's degree in Mathematics and Applied Mathematics from Beijing University of Posts and Telecommunications, in 2020. She is working toward the master's degree in the Key Laboratory of Trustworthy Distributed Computing and Services, Beijing University of Posts and Telecommunications, Ministry of Education, China. Her research interests include data mining and machine learning.
\end{IEEEbiography}

\begin{IEEEbiography}[{\includegraphics[width=1in,height=1.25in,clip,keepaspectratio]{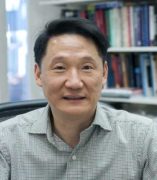}}]{Philip S. Yu} received the B.S. Degree in E.E. from National Taiwan University, the M.S. and Ph.D. degrees in E.E. from Stanford University, and the M.B.A. degree from New York University. He is a Distinguished Professor in Computer Science at the University of Illinois at Chicago and also holds the Wexler Chair in Information Technology. Before joining UIC, Dr. Yu was with IBM, where he was manager of the Software Tools and Techniques department at the Watson Research Center. His research interest is on big data, including data mining, data stream, database and privacy. He has published more than 1,500 papers in refereed journals and conferences. He holds or has applied for more than 300 US patents. Dr. Yu is a Fellow of the ACM and the IEEE. Dr. Yu is the recipient of ACM SIGKDD 2016 Innovation Award for his influential research and scientific contributions on mining, fusion and anonymization of big data. He also received the VLDB 2022 Test of Time Award, ACM SIGSPATIAL 2021 10-year Impact Award, and the EDBT 2014 Test of Time Award. He was the Editor-in-Chiefs of ACM Transactions on Knowledge Discovery from Data (2011-2017) and IEEE Transactions on Knowledge and Data Engineering (2001-2004).
\end{IEEEbiography}







\end{document}